\documentclass[journal,draftclsnofoot,onecolumn,12pt]{IEEEtran}

\usepackage[utf8]{inputenc}
\usepackage[english]{babel}

\ifCLASSINFOpdf
\else
\fi
\usepackage{amsmath,amsfonts,amssymb,stackrel,bbm,dsfont,bm,float}
\usepackage{enumitem}
\usepackage[section]{placeins}
\usepackage{algorithm}[2] 
\usepackage[noend]{algpseudocode}
\usepackage{cite} 
\usepackage{microtype,graphicx,booktabs,caption,subcaption,hyperref,xcolor}
\graphicspath{./figures}

\newtheorem{remark}{Remark}

\usepackage{graphicx}
\DeclareGraphicsExtensions{.pdf}

\begin{document}

\pagenumbering{arabic} 

\title{Deep Reinforcement Learning for Resource Constrained Multiclass Scheduling in Wireless Networks}

\author{Apostolos Avranas, 
        Philippe Ciblat,~\IEEEmembership{Senior Member, IEEE,}\\
        and~Marios Kountouris,~\IEEEmembership{Senior Member,~IEEE.}
\thanks{A. Avranas and M. Kountouris are with the Communication Systems department, EURECOM, F-06904 Sophia-Antipolis, France. Emails: apostolos.avranas@eurecom.fr, marios.kountouris@eurecom.fr. P. Ciblat is with LTCI, Telecom Paris, Institut Polytechnique de Paris, F-91120 Palaiseau, France. Email: philippe.ciblat@telecom-paris.fr}}

\maketitle
\begin{abstract}
The problem of resource constrained scheduling in a dynamic and heterogeneous wireless setting is considered here. In our setup, the available limited bandwidth resources are allocated in order to serve randomly arriving service demands, which in turn belong to different classes in terms of payload data requirement, delay tolerance, and importance/priority. In addition to heterogeneous traffic, another major challenge stems from random service rates due to time-varying wireless communication channels. Various approaches for scheduling and resource allocation can be used, ranging from simple greedy heuristics and constrained optimization to combinatorics. Those methods are tailored to specific network or application configuration and are usually suboptimal. To this purpose, we resort to deep reinforcement learning (DRL) and propose a distributional Deep Deterministic Policy Gradient (DDPG) algorithm combined with Deep Sets to tackle the aforementioned problem. Furthermore, we present a novel way to use a Dueling Network, which leads to further performance improvement. Our proposed algorithm is tested on both synthetic and real data, showing consistent gains against state-of-the-art conventional methods from combinatorics, optimization, and scheduling metrics.
\end{abstract}


\IEEEpeerreviewmaketitle


\section{Introduction}
Scheduling and resource allocation are two relevant yet challenging problems with a plethora of practical applications in various fields. For instance, in computing systems, computational processes have to be efficiently arranged and planned for the server to handle them; in management, each person is assigned a set of jobs for completion; in logistics, packages have to be carefully matched to each truck. Since resources, e.g., central processing units (CPUs), workers, trucks, etc., are limited, they have to be shared efficiently among different tasks and requests so as to optimize the system performance. 
Optimal resource allocation, together with the associated scheduling task, is one of the main challenges and requirements for the design of communication networks. How efficiently the available resources (e.g., subbands, timeslots, beams, transmit power, etc.) are managed has a direct impact on the communication system performance.

In this paper, we investigate the problem of scheduling and resource allocation in wireless networks. A base station (BS) sends data traffic to mobile users, which have different application-dependent Quality of Service (QoS) requirements. We consider applications that require delivery of large amounts of data without any strict deadline, as well as time-sensitive or mission-critical ones involving low payload packets that have to be reliably received within a stringent latency constraint. The increased heterogeneity in users' traffic and the diverse service requirements substantially complicate the provisioning of high fidelity, personalized service with QoS guarantees. The objective of this work is to design a generic architecture and efficient algorithms, which take as inputs the specific constraints of the traffic/service class where each user belongs to and as outputs the set of users to serve, as well as the allocated resources and timeslots, as means to maximize the number of satisfied users.
    
The considered problem here is hard to solve due to several major technical challenges. First, with the exception of very few special cases, there is no simple closed-form expression for the problem and {\it a fortiori} for its analytical solution. Second, optimization algorithms that solve the problem have to be computationally efficient and implementable in large-scale wireless networks. Applying optimal methods from combinatorial optimization, such as branch and bound algorithm \cite{BranchAndCutIntro}, results in solutions exhibiting prohibitively high  computational complexity and being hard or impossible to meaningfully scale with the number of active users. Other existing approaches relying heuristics, approximations, or relaxations, provide suboptimal solutions, which seem to work satisfactorily in specific scenarios but fail to perform close to optimal in general cases and with large number of users. Moreover, the proliferation of new use cases makes the problem of efficient and scalable scheduling and resource allocation more intricate. This will be exacerbated with the advent of the emerging mobile systems (Beyond 5G/6G), which will involve high-dimensional optimization domains, various application scenarios, as well as heterogeneous, often conflicting, QoS requirements. This motivates the quest for alternative methods. 

In this paper, we propose to resort to Deep Reinforcement Learning (DRL) for efficient and scalable resource allocation. DRL has recently attracted much attention for providing very promising results in complex problems obeying strict game rules (e.g., Atari, Chess, Go \cite{AtariNature,AlphaZero,AlphaGoDeepmind}) or physical laws (robotics and physics-related tasks \cite{RL_inRobotics,continuousControlDPG}). In cloud service provision, DRL has been used to schedule incoming tasks to servers according to their heterogeneous CPU and memory requirements \cite{DRLcloud}. DRL approaches have recently shown interesting gains in wireless communication systems \cite{cite3,cite1,cite2,9120241,MultiAgent2,DRLComm1,DRLComm2,DRLComm3,DRLComm4}. In contrast to most prior work and to harness the high level of stochasticity, we consider distributional DRL \cite{JaquetteFirstDistr,IQNdistributionalRL,QR-DQN} as a means to obtain richer representations of the environment thus better solutions. Furthermore, we leverage (i) techniques such as noisy networks for better explorations \cite{NoisyNets}; (ii) architectures such as dueling networks \cite{DuelingDRL} for improved stability of the trained models; and (iii) deep sets \cite{DeepSets} for simplifying and improving neural network models when permutation invariance properties apply. We combine these three ingredients with a deep deterministic policy gradient method \cite{DeepDPG} to propose a highly efficient general architecture and scheduling/resource allocation algorithm.
In a setup similar to ours and Nokia's challenge \cite{NOKIAchallenge}, deep deterministic policy gradient is used to allocate the bandwidth to incoming data traffic in \cite{DDPG_scheduling_withReal}. Nevertheless, unlike our work, \cite{DDPG_scheduling_withReal} considers only full channel state information (CSI), a single traffic class, and only few users (typically less than $15$). In \cite{GraphNN_general}, Graph Neural Networks, a technique similar to Deep Sets, are used to increase the number of users, but they do not consider traffic of users. Initial attempts to solve the problem of scheduling traffic for users with heterogeneous performance requirements can be found in \cite{eMBB_URLLC_drl}, considering though only full CSI and a limited number of users.

In the context of using DRL for revisiting the problem of heterogeneous multiclass scheduling and dynamic resource allocation in wireless communication networks, our main contributions can be summarized as follows:
\begin{itemize}
    \item We develop a neural network with two crucial architectural choices facilitating a stable training even in the case of high traffic from a very large number of users. First, we leverage \textit{Deep Sets} \cite{DeepDPG} as a means to exploit the permutation equivariance property of the problem and drastically reduce the number of necessary parameters. Second, we introduce a \textit{user normalization} trick capturing the attribute of the problem that the available bandwidth resources are limited. We show that without those crucial steps, the performance plummet.
    \item We further improve the performance using distributional DRL \cite{QR-DQN} and reward scaling as implemented in \cite{ImplementationMattersDeepRL_RewardScaling}. Finally, we get additional gains by adapting the idea of dueling networks \cite{DuelingDRL} used in Deep Q-Networks (DQN) to distributional RL by modifying the output to represent the distribution of the return of the agent's action.
    \item We demonstrate that our DRL proposal can easily be implemented with minor changes in both  extremal cases in terms of wireless channel knowledge, namely full CSI and no CSI.
    \item To compare our DRL solution, we design strong baselines:
    \begin{itemize}
        \item In the \textit{full CSI} case, the scheduling step is solved in a myopically optimal way by reformulating it as a knapsack problem. The DRL scheduler outperforms baseline schemes in the sense that it reaches the same performance while requiring  $13\%$ less power and bandwidth. Furthermore, we devise a baseline operating as an oracle knowing all future traffic characteristics. The oracle finds the optimal resource allocation policy via Integer Linear Programming (ILP) and constitutes an upper bound. Our experimental results show that the proposed DRL scheduler operates close to the upper bound.
        \item In the \textit{no CSI} case, the model-based baseline actually requires access to the statistics of the problem and uses them to cast the scheduling problem as an optimization one. The baseline scheme employs the Frank-Wolfe (FW) algorithm, which guarantees that the solution is a local optimum. Our DRL scheme significantly outperforms Frank-Wolfe, supporting our hypothesis that the more complicated the communication system is with unknown variables affecting it, the higher gains may be yield using a DRL-based model-free method.
    \end{itemize} 
\end{itemize}

The paper is organized as follows: in Section \ref{sec:model}, we introduce the system model including the channel and traffic model. In Section \ref{sec:problem}, we formulate the optimization problem and Section \ref{sec:drl} is devoted to the main contribution of the paper, that of the design a new DRL scheduler for heterogeneous multiclass traffic. In Section \ref{sec:baseline}, baseline algorithms, for performance comparison, are presented. In Section \ref{sec:experiments}, we provide experimental results with both synthetic and real data, and Section \ref{sec:ccl} concludes the paper.

\section{System Model}\label{sec:model}
    
\subsection{Network and channel model}
We consider the downlink of a communication system, in which a BS serves multiple users by sending data over a wireless random time-varying channel. Users are uniformly distributed within two concentric rings of radii $\mathrm{d}_{min}$ and $\mathrm{d}_{max} > \mathrm{d}_{min}$. Therefore, the distance of a user $u$ from the BS is a random variable with probability density function (PDF) $f_\mathrm{d}(\mathrm{d}_u)=\frac{2\mathrm{d}_u}{\mathrm{d}_{max}^2-\mathrm{d}_{min}^2}, \mathrm{d}_u\in[\mathrm{d}_{min},\mathrm{d}_{max}]$. We assume that mobility is not very high, such that BS-user distances remain constant during the time interval users remain active.

Orthogonal frequency bands are assigned to simultaneously served users, hence there is no interference among them. Users experience frequency flat fading, i.e., the channel gain of a user remains constant during a time slot and throughout all available frequency bands assigned to. Let a user $u$ that has entered the system at time $t_0$. Its channel gain at time $t$ is given by $ g_{u,t}=\frac{C_{pl}|h_{u,t}|^2}{\sigma_N^2} \mathrm{d}_u^{-n_{pl}}$, where $n_{pl}$ denotes the pathloss exponent, $C_{pl}$ is a constant accounting for constant losses, and $\sigma_N^2$ is the noise power spectrum density. The small-scale fading $h_{u,t}$ evolves over time according to the following Gauss-Markov model
    \begin{eqnarray}
    h_{u,t} = \rho h_{u,t-1} + N \label{eq:Small Scale Fading Markov}
    \end{eqnarray}
    where $h_{u,t_0} \sim \mathcal{CN}(0,1)$ (circular complex normal distribution with zero mean and unit variance), and $N\sim \mathcal{CN}(0,1-\rho ^2), \  t>t_0$. 
    The parameter $\rho=J_0(2\pi f_d T_{slot})\in[0,1]$ \cite{MarkovChannelModel} determines the time correlation of the channel, with $J_0(\cdot)$ denoting the zeroth-order Bessel function of the first kind, $f_d$ the maximum Doppler frequency (determined by the user mobility), and $T_{slot}$ the slot duration. 
    If $\rho=0$ (high mobility), a user experiences an independent realization of the fading distribution at each time slot (i.i.d. block fading). If $\rho=1$ (no mobility), channel attenuation is constant throughout the user's lifespan (no small-scale fading fading). 
    
    We consider the following two cases for the channel state information (CSI): (i) \textit{full-CSI}, in which $h_{u,t_c}$ and the users' locations (and so $\mathrm{d}_u$) are perfectly known at the BS for time $t_c$, thus enabling accurate estimation of the exact resources each user requires; (ii) \textit{no-CSI}, in which the scheduler is completely channel-agnostic, both in terms of instantaneous fading realization and long-term channel statistics. In case of unsuccessful and/or erroneous data reception, a simple retransmission protocol (Type-I Hybrid Automatic Repeat Request (HARQ)) is employed. A packet is discarded whenever the user fails to correctly decode it (no buffering at the receiver side) and the BS will attempt to send it again in some subsequent slot.
    
    \begin{remark}\label{rem:1}
    For a non-trivial implementation of the Frank-Wolfe (FW) algorithm, which serves as a baseline for comparison in the the no-CSI case, we need to consider some kind of CSI. For that, we consider the case of \textit{statistical CSI}, where the scheduler knows the statistics of the users' channels and locations. Our proposed DRL algorithm will always operate under full absence of CSI, since all statistics can effectively be learned through the training phase. 
    \end{remark}

\subsection{Traffic model}
\label{Traffic model}
We consider a generic yet tractable traffic model, in which users with diverse data and latency requirements arrive and depart from the system. There is a set of service classes $\mathcal{C}$ to which a user entering the system belongs to with probability $p_c$. Each user in class $c\in \mathcal{C}$ is characterized by the tuple $(D_c, L_c, \alpha_c)$ as follows: 
\begin{itemize}
    \item{Data size $D_c$}: the number of information bits requested by a user belonging to class $c$.
    \item{Maximum Latency $L_c$}: the maximum number of time slots within which the user has to be satisfied, i.e., to successfully receive its data packets of size $D_c$.
    \item{Importance $\alpha_c$}: an index allowing the scheduler to prioritize certain service classes, e.g., users with privileged contracts (e.g., high-value Service-Level Agreement (SLA)) may demand better service and higher reliability. 
\end{itemize}
We assume that a maximum number of users $K$ can coexist per time slot and that a new user may arrive only after the departure of a user that exceeded the maximum time allowed to remain in the system. That way, the scheduling decisions do not influence the arrival process. For example, if a user arrives at time $t_0=1$, belonging to class $c\in\mathcal{C}$ with $L_c = 4$, then even if it successfully receives its requested packet of size $D_c$ at $t=1$, a new  arrival may randomly be generated only at time $t=t_0+L_c=5$ and afterwards. The rationale behind adopting this model is as follows. If a new arrival is generated right after a previous user is satisfied (in the example at time $t=2$), then the traffic load is affected by the scheduler performance. The faster the scheduler serves the users, the more arrivals occurs. In contrast, in our model, the arrival process and its statistics remain uninfluenced by the scheduling decisions and the available resources.  
Therefore, at every time slot, the set of users $U_t$ ($|U_t|\leq K$) contains all users waiting to receive their requested data while remaining within their latency constraint. Finally, to ensure random inter-arrival times, we assert that the probability $p_{null} = 1 - \sum_{c\in \mathcal{C}}p_c$ is positive , i.e., $p_{null}>0$, leaving a probability that no user appears in a time slot. 
    
\subsection{Service Rate}
\label{sec:Rate model}
The service rate is measured using Shannon rate expression assuming capacity-achieving codes. The achievable service rate of user $u$ at time $t$ is equal to $w_{u,t}\mathcal{R}_{u,t}$, where 
    $\mathcal{R}_{u,t} = \log_2(1+g_{u,t}\mathrm{P}_{u,t})=\log_2(1+\kappa_u|h_{u,t}|^2)$ (bit/s/Hz), with $\mathrm{P}_{u,t}$ denoting the transmit energy per symbol (channel use), $w_{u,t}$ the assigned bandwidth (in Hz), and $\kappa_u =\frac{C_{pl}}{\sigma_N^2} \mathrm{d}_u^{-n_{pl}}$.
    Let a user at distance $d_u$ from the BS, belonging to class $c\in \mathcal{C}$, is served at time $t_u$ with resources $(w_{u,t},\mathrm{P}_{u,t})$. The probability of unsuccessful transmission is given by 
    \begin{eqnarray}
    P_{u}^{\textrm{fail}}(w_{u,t},\mathrm{P}_{u,t};\mathrm{d}_u) = \mathbbm{P}(w_{u,t}\mathcal{R}_{u,t}  < D_u|\mathrm{d}_u)  = \mathbbm{P}(|h_{u,t}|^2< \zeta_{u,t} \mathrm{d}_u^{n_{pl}}) = 1 - {\displaystyle e^{-\zeta_{u,t} \mathrm{d}_u^{n_{pl} }}}
    \label{eq:FailureProb_u_t_given_dist}
    \end{eqnarray}
    where $\zeta_{u,t}{=}\displaystyle \frac{\sigma_N^2(2^{D_u/w_{u,t}}-1)}{C_{pl}\mathrm{P}_{u,t}}$. 
    If $d_u$ is not known to the scheduler, we have 
    \begin{eqnarray}
    P_{u}^{\textrm{fail}}(w_{u,t},\mathrm{P}_{u,t}) &=& \mathbbm{P}(w_{u,t}\mathcal{R}_{u,t}< D_u) = \int_{\mathrm{d}_{min}}^{\mathrm{d}_{max}}  P_{u}^{\textrm{fail}}(w_{u,t},\mathrm{P}_{u,t};\mathrm{d})  f_{\mathrm{d}}(\mathrm{d} )d\mathrm{d} \nonumber\\
    &=& 1 -{\displaystyle \frac{\Gamma(\frac{2}{n_{pl}},\zeta_{u,t}\mathrm{d}_{min}^{n_{pl}}){-}\Gamma(\frac{2}{n_{pl}},\zeta_{u,t}\mathrm{d}_{max}^{n_{pl}})}   {n_{pl}\zeta_{u,t}^{2/n_{pl}}(\mathrm{d}_{max}^2-\mathrm{d}_{min}^2)/2}}
    \label{eq:FailureProb_u_t}
    \end{eqnarray}
    where $\Gamma(s,x) = \int_x^{\infty} t^{s-1}\,\mathrm{e}^{-t}\,{\rm d}t$ is the upper incomplete gamma function. For exposition convenience, we overload notation by allowing $u$ in $D_u,L_u,\alpha_u$ to denote either a class $u$ or a user $u$ belonging to a class with those characteristics.


\section{Problem Statement}\label{sec:problem}
We consider the problem of heterogeneous scheduling and resource allocation, which involves a BS handling a set of randomly arriving service requests belonging to different classes with heterogeneous requirements. Each class defines the requirements and the expected Quality of Service (QoS) guarantees for its users. Observing this time-varying set of heterogeneous requests, the objective of the scheduler at each time slot is two-fold: (i) carefully select which subset of user requests to satisfy, and (ii) allocate the finite resources amongst the selected user requests. The performance metric to maximize is the long-term importance-based weighted sum of successfully satisfied requests. A request is considered to be satisfied whenever the user has received the requested data within the maximum tolerable latency specified by its service class.

The scheduling problem at hand can be formulated  as a Markov Decision Process (MDP) \cite{bellmanMDP} $(\mathcal{S},\mathcal{A}, R,$ $P, \gamma)$, where $\mathcal{S}$ is the state space of the environment and $\mathcal{A}$ is the action space, i.e., the set of all feasible allocations in our case. After action $a_t\in \mathcal{A}$ at state  $s_t\in\mathcal{S}$, a reward $r_t\sim R(\cdot |s_t,a_t)$ is obtained and the next state follows the probability $s_{t+1} \sim P(\cdot |s_t,a_t)$. The discount factor is $\gamma\in[0,1)$. Under a fixed policy $\pi: \mathcal{S} \rightarrow \mathcal{A}$ determining  the action at each time step, the \textit{return} is defined as the random variable 
\begin{equation}\label{eq:ReturnRandomVariable}
    Z^\pi_t = \sum_{i=0}^\infty \gamma^{i} r_{t+i} 
\end{equation} 
which represents the discounted sum of rewards when a trajectory of states is taken following $\pi$. Ideally, the aim is to find the optimal policy $\pi^\star$ that maximizes the mean reward $\mathbbm{E}_\pi[Z^\pi_t]$. 
            
At each time step $t$, a set of users $u\in U_t$ is waiting for service, where each user therein belongs to a class $c_u\in\mathcal{C}$ described by $(D_c, L_c, \alpha_c)$. After $L_c$ time steps a new user belonging to class $c$ might arrive with probability $p_c$. Throughout its ``lifespan'' $t\in [t_0,t_0{+}L_c{-}1]$, $w_{u,t}$ indicates the amount of given resources. If at any time $t$, $w_{u,t}>D_u/\mathcal{R}_{u,t}$ then user $u$ is satisfied. Since resources are limited (finite), $\sum_{u\in U_t}w_{u,t}\leq W, \quad \forall t$, no more than $W$ resources in total can be spent per time slot. 
Summing up,
\begin{itemize}
\item[] State: $s_t = \{\forall u \in U_t: c_u, \mathcal{R}_{u,t}, l_{u,t}\}$
\item[] Action: $a_t = \{\forall u \in U_t: w_{u,t}\}$
\item[] Reward: $\displaystyle r_t = \sum_{u\in U_t} \alpha_u \mathds{1}\{w_{u,t} \mathcal{R}_{u,t}>D_u \}$
\end{itemize}
where $l_{u,t}\leq L_u$ is the remaining number of time slots  within which user $u$ (i.e. $u\in U_t$) expects to successfully receive its packet  and $\mathds{1}\{\cdot\}$ denotes the indicator function. 
Note that knowing the class $c_u$ to which user $u$ belongs, implies knowing the requirements  $(D_u,L_u,\alpha_u)$. An inherent attribute of this MDP is the \textit{permutation equivariance} of an optimal policy, meaning that if we permute the indexing of the users, then permuting likewise the allocation of the resources retains the performance of the policy. For that, in our DRL approach, we only consider permutation equivariant policies, and as a consequence we need a permutation \textit{invariant function} to evaluate and train the policy. 
            
In this work, we focus on bandwidth allocation, assuming a fixed amount of energy spent per channel use and no power adaptation, i.e., $\mathrm{P}_{u,t}=\mathrm{P}, \forall u,t$. 
Specifically, for total bandwidth $W$, the scheduler aims at finding the $(w_{u_1,t},w_{u_2,t},...)\in \mathbbm{R}_{\geq 0}^{|U_t|}$ with $u_1,u_2,...\in U_t $ and $\sum_{u\in U_t}w_{u,t}\leq W, \forall t$, so as to maximize the accumulated reward for every satisfied user over a finite time horizon. The expected reward is described by the following objective ``gain-function"
    \begin{eqnarray}
        G = \sum_{t}\sum_{u\in U_t} \alpha_u \mathds{1}\{ w_{u,t}\mathcal{R}_{u,t}>D_u \}.\label{eq:GainFunc}   
    \end{eqnarray}
We stress out that a user $u$ remains on the set $U_t$ for a time interval less or equal to the maximum acceptable latency $L_u$. If not satisfied within that interval, then it does not contribute positively to the objective $G$.
    
Note that $\mathcal{R}_{u,t}$ satisfies the Markov property since $h_{u,t}$ follows a Markov model. Under full CSI, the agent (here the BS) fully observes the state $s_t$, while in the no-CSI case, $h_{u,t}$ is unknown resulting in a Partially Observable MDP (POMDP) \cite{PartiallyObservableMDP}. One way to transform a POMDP into a MDP is by substituting the states with the ``belief" of the states  \cite{POMDB_planning}. Another way is to use the complete history $\{o_0,a_0,o_1,a_1,\cdots,a_{t-1},o_{t-1}\}$, with $o_t\subset s_t$ being the agent's observation. Notice that only the most recent part is relevant as users that have already left the system do not affect the way the channels of the current users evolve or the generation of future users or in general the current and future system dynamics. Therefore, we can safely consider the scheduling and allocation history of only the current users. Specifically, if ${\bm{w}}_{u,t}=(w_{u,t_0},$ $w_{u,t_0+1}, ... ,w_{u,t})$ is the scheduling history of user $u$ then the input of the agent is $ \{\forall u \in U_t: D_u,L_u,a_u, \kappa_u, l_{u,t},{\bm{w}}_{u,t}\}$.

\section{Proposed Deep Reinforcement Learning Architecture}\label{sec:drl}
In this section, we propose a novel DRL architecture as a means to solve the aforementioned multiclass scheduling and resource allocation problem. Despite the highly challenging dynamics and stochasticity (wireless channel and heterogeneous traffic), we show that DRL can provide performance gains although it is impossible to accurately predict the number of users, their service demands, and their channel/link characteristics even after few steps.

\subsection{Policy Network}
\label{sec:PolicyNet}
Our objective is to build a scheduler that can handle a large number of users $K$, even in the order of hundreds. Moreover, we require that our method works in both full CSI and no CSI cases with minor - if any - modifications. A widely used approach is Deep Q-learning Network (DQN). However, it is not feasible to employ DQN in our case since it needs a Neural Network (NN) architecture with a number of outputs equal to the number of possible actions and the action space is extremely large (in statistical CSI it is even infinitely large). For that, we resort to a Deep Deterministic Policy Gradient method \cite{DeepDPG}, which trains a policy $\pi_\theta: \mathcal{S}\rightarrow \mathcal{A}$ modeled as a NN with parameters $\theta$. 
        
    If at time $t$ on state $s_t$ the action $a_t$ is taken followed by the policy $\pi$, then the return using \eqref{eq:ReturnRandomVariable} is given by 
    \begin{equation}
        \label{eq:Return(st,at)}
        Z^\pi (s_t,a_t)= r_{t}+\gamma Z^\pi_{t+1}, \text{with} \ r_{t}\sim R(\cdot |s_{t},a_{t}).
    \end{equation}
     Note that if even at $t$ the action $a_t$ comes from policy $\pi$, then $Z^\pi (s_t,a_t=\pi(s_t))=Z^\pi_{t} $. Let the expected return be 
    \begin{equation}
        \label{eq:Q-function}
        Q^\pi(s_{t},a_{t})=\mathbbm{E}[Z^\pi (s_{t},a_{t})].
    \end{equation}
    Then, the objective of the agent is to maximize 
    \begin{equation} 
        \label{eq:ActorObjective}
        J(\theta)=\mathbbm{E}_{s_{t_0}\sim \mathrm{p}_{t_0} }[Q^{\pi_\theta}( s_{t_0},\pi_\theta(s_{t_0}))],
    \end{equation}
    with $\mathrm{p}_{t_0}$ being the probability of the initial state $s_{t_0}$ at time $t_0$. The gradient can be written \cite{DeterministicPG}
    \begin{equation}
        \label{eq:direvativeActor}
        \nabla_\theta J(\theta) = \mathbbm{E}_{s_{t_0}\sim \mathrm{p}_{t_0},s\sim\rho^{\pi_\theta}_{s_{t_0}}}[\nabla_\theta \pi_\theta(s)\nabla_a Q^{\pi_\theta}(s,a)|_{a=\pi_\theta(s)}], 
    \end{equation}
    with $\rho^{\pi_\theta}_{s_{t_0}}$ being the discounted state (improper) distribution defined as $\rho^{\pi_\theta}_{s_{t_0}}(s)= \sum_{i=0}^\infty \gamma^i \mathbbm{P}(s_{t+i}=s|s_{t_0},\pi_\theta)$. In practice $\rho^{\pi_\theta}_{s_{t_0}}$ is approximated by the (proper) distribution $\varrho^{\pi_\theta}_{s_{t_0}}(s) := \sum_{i=0}^\infty \mathbbm{P}(s_{t+i} = s|s_{t_0},\pi_\theta)$. To compute the gradient, the function $Q^{\pi_\theta}(s,a)$ is needed, which is approximated by another NN $Q_\psi(s,a)$, named \textit{value network}, described in the next subsection. 
                
We now explain the architecture of the model $\pi_\theta$. 
    
\subsubsection{Deep Sets} 
As discussed in Section \ref{sec:problem}, we aim for a policy that falls in the realm of permutation equivariant functions (i.e., permuting the users should only result in permuting likewise the resource allocation). In \cite{DeepSets}, necessary and sufficient conditions are shown for permutation equivariance in neural networks; their proposed structure called Deep Sets is adopted here. At first, the characteristics (or \textit{features} as commonly termed in the machine learning literature) $F_i\in \mathbbm{R}^{N_{u} }, i\in\{1,\cdots K\}$ of each user are processed individually by the same function $ \phi_{user}:\mathbbm{R}^{N_u}\rightarrow \mathbbm{R}^{H_u}$ modeled as a two layer fully connected neural network. Then, the outputs of $\phi_{user}$ that corresponds to the new characteristics per user are aggregated with the permutation equivariant $f_\sigma:\mathbbm{R}^{K\times H} \rightarrow \mathbbm{R}^{K\times H'}$ of $H$ (resp. $H'$) input (resp. output) characteristics:
\begin{equation}
            \label{eq:DeepSets}
            f_\sigma (x) = \sigma\left(x\Lambda + \frac{1}{K}\mathbf{1}\mathbf{1}^\intercal x \Gamma\right), \qquad \Lambda,\Gamma\in\mathbbm{R}^{H\times H'}
\end{equation}
where $\mathbf{1} = [ 1, \cdots,1]\in \mathbbm{R}^K$ and $\sigma(\cdot)$ is an element-wise nonlinear function. We stack two of those, one $f_{\mathrm{relu}}:\mathbbm{R}^{K\times H_u} \rightarrow \mathbbm{R}^{K\times H_u'}$ with $\sigma(\cdot)$ being the $\mathrm{relu}(x) = \max(0,x)$ and a second $f_{\mathrm{linear}}:\mathbbm{R}^{K\times H_u'} \rightarrow \mathbbm{R}^{K\times 1}$ without any nonlinearity $\sigma(\cdot)$. In addition to preserving  the desirable permutation equivariance property, this structure also brings a significant parameter reduction. The number of parameters of Deep Sets contained in $\Lambda,\Gamma$ do not depend on the number of users $K$. Therefore, any increase in $K$ does not necessitate additional parameters, which could lead to a much bigger network, prone to overfitting.

\subsubsection{Output} 
The activation function for the last layer of the policy network is a smooth approximation of $\mathrm{relu}(x)$, namely $\mathrm{softplus}(x) = \log(1+e^x)$ restricting the output $\bm{y} \in \mathbbm{R}^K$ to be positive. After that, depending on the existence of CSI, there are two ways of performing the allocation. For full CSI, the bandwidth required per user is accurately known. Therefore, we only need a binary decision per user (to serve or not), which will ruin though the differentiability of the policy, a mandatory property for DDPG to work. For that, we interpret the output $\bm{y}$ as a continuous relaxation of the binary problem. Specifically, $\bm{y}$ is the assignment to each user of a  ``value'' per resources which after being multiplied by the number of resources the user requires, a user ranking is obtained. Then, the scheduler satisfies as many of the most ``valuable'' (highest rank) users as possible subject to available resources. Therefore, in full CSI, $\bm{y}$ semantically denotes how advantageous the policy believes is to allocate resource to each user. 
On the contrary, in the no CSI case, the action is not binary but continuous since the scheduler has to decide on the portion of the available resources each user takes. To ensure that $\bm{y}$ has the valid form of portions (i.e., positive and adding up to one) we just divide by the sum, $\bm{y}\rightarrow \frac{\bm{y}}{||\bm{y}||_1}$ (with $||\cdot||_1$ being the $\mathrm{\ell}_1$ norm)\footnote{Instead of dividing by the $\mathrm{\ell}_1$ norm, we also considered the $\mathrm{softmax}(\bm{y})$, which seemed a good choice as it also provides positive outputs adding up to one. Nevertheless, this approach led to poor performance because no matter how much the number of users is increased, the policy insists on evaluating as advantageous to serve only a very small number of users. This makes sense since the softmax function is a smooth approximation of $\mathrm{argmax}$, hence focusing on finding the one most advantageous user to be served.}. This discrepancy in the output process is the only minor difference in the considered model between full CSI and no CSI. 
        
\begin{figure}
\centering
\includegraphics[width=0.45\textwidth]{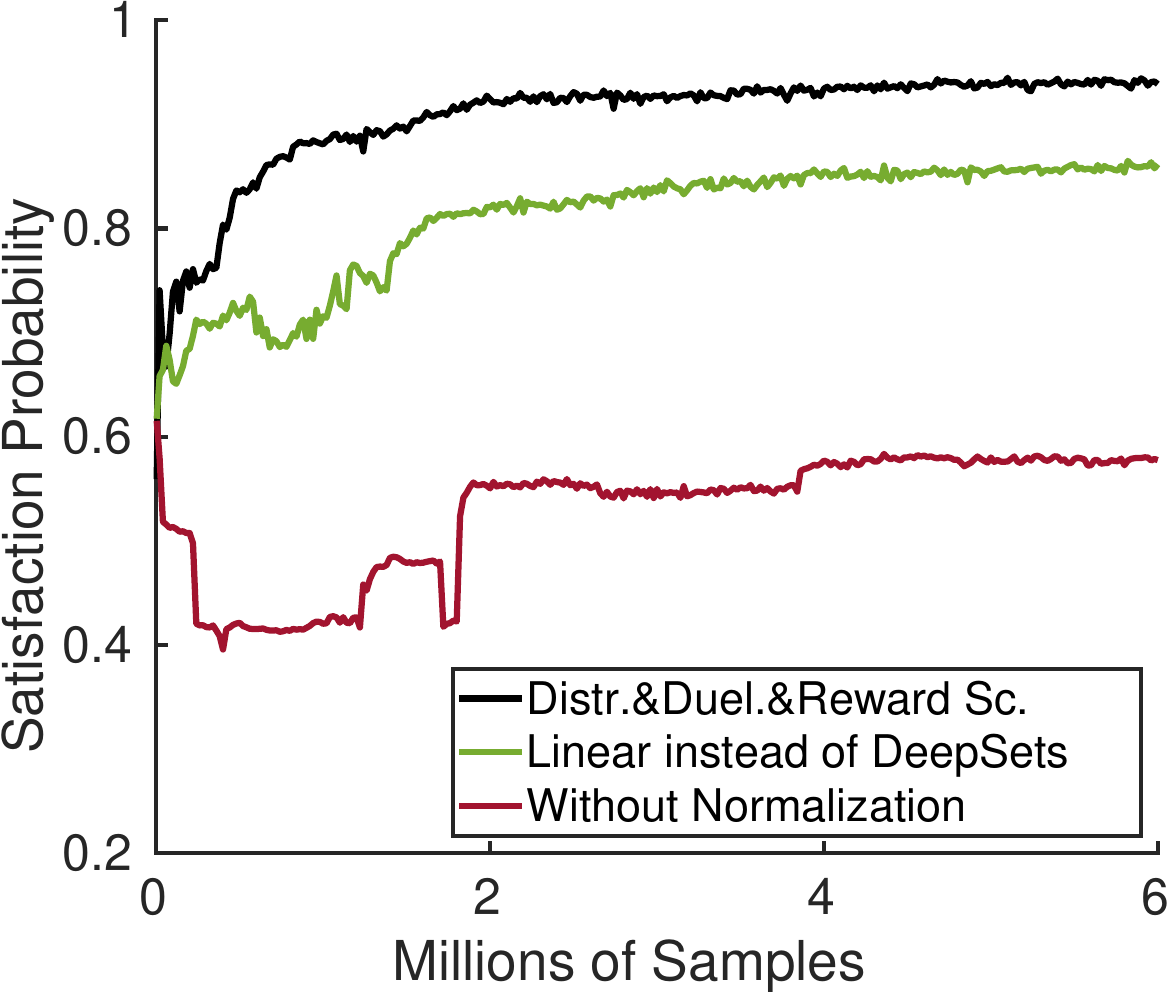}
\caption{We conducted five experiments (with different seeds) for no CSI using the traffic model of Table \ref{tab:Equal Users}, a maximum number of users $K=75$, $\rho=0$, and resources (total bandwidth) $W=5$ MHz. We depict here the average probability a user to be satisfied over those five experiments to carry an ablation study on the importance of the deep sets and user normalization step.}
\label{fig:WithoutDeepSetsORNormal}
\end{figure}
        
\subsubsection{User normalization} 
Before the final nonlinearity of $\mathrm{softplus}(x) = \log(1+e^x)$, as seen in Figure \ref{fig: NetworkArchitecture}, there is the crucial ``user normalization'' step  $\bm{x}\rightarrow \frac{\bm{x}-\mathbbm{E}[\bm{x}]}{||\bm{x}||_2}, \bm{x} \in \mathbbm{R}^K$ (with $||\cdot||_2$ denoting the $\mathrm{\ell}_2$ norm). Consider first the full CSI case. Without that step, the value network would perceive that the higher the ``value'' per resource assigned to a user, the more probable is for that user to get resources (and thus to be satisfied and receive reward). Unfortunately, this leads to a pointless interminable increase of every user's ``value''. What matters here is not the actual ``value'' of a user but how large this is relative to the rest of the users. To bring the notion of limited total resources, the ``user normalization" subtracts from the value of each user the mean of all the users value. Hence, whenever the algorithm pushes the value of a single user to increase, the values of the rest decrease. In the no CSI case, there is an additional benefit. Since in the following step there is the operation $\bm{y}\rightarrow\frac{\bm{y}}{||\bm{y}||_1}$ so as the output to signify portions (of the total bandwidth), performing previously the normalization step (dividing by $||\bm{x}||_2$) helps keeping the denominator $||\bm{y}||_1$ stable. 
                    
In Figure \ref{fig:WithoutDeepSetsORNormal} we show the significance of choosing the right architecture. It is clearly observed that \emph{if either all Deeps Sets (in both policy and value network) are substituted by the most common choice of linear blocks or the user normalization step is removed, the performance degrade substantially.}
                
\subsubsection{Exploration} 
Since the action $a_t$ has to satisfy specific properties, such as positiveness and summing up to one for the no CSI case, the common approach of adding noise on the actions becomes rather cumbersome. An easy way out is through noisy networks \cite{NoisyNets}, which introduce noise to the weights of a layer, resulting to change decisions for the policy network. The original approach considers the variance of the added noise to be learnable. Here, we instead keep it constant since this provides better results. With probability $P_{explore}$ we add noise to the parameters of $\phi_{users}$, resulting to alter output features per user and therefore the policy outputs a different allocation. Specifically, if $\theta_{\phi_{users}}$ are the parameters of $\phi_{users}$, then they are distorted as $\theta_{\phi_{users}}(1+\sigma_{explore}\epsilon)$ with $\epsilon$ being normally distributed with zero mean and unit standard deviation and $\sigma_{explore}$ being a constant.

    \subsection{Value Network}
    As mentioned previously, $Q^{\pi_\theta}(s,a)$ is used for computing the gradient of the objective function described in \eqref{eq:ActorObjective}. Since this is intractable to compute, a neural network, named value network, is used to approximate it. We compare three ways of employing the value network.
            
        \subsubsection{DDPG} 
        At first, the common approach of DDPG is considered, which uses the Bellman operator
        \begin{equation}\label{eq:Common Bellman}
            \mathcal{T}^{\pi}Q(s,a)  = \mathbbm{E}_{r\sim R(s,a),s'\sim  P(s,a)}[r+\gamma Q(s',\pi (s))] 
        \end{equation}
        to minimize the temporal difference error, i.e., the difference between before and after applying the Bellman operator. This leads to the minimization of the loss
        \begin{equation}
            \mathcal{L}_2(\psi) =\mathbbm{E}_{s_{t_0}\sim \mathrm{p}_{t_0},s\sim \rho^{\pi_\theta}_{s_{t_0}} }[(Q_\psi(s,a)-\mathcal{T}^{\pi_{\theta'}}Q_{\psi '}(s,a))^2]
        \end{equation}
        where $(\pi_{\theta'},Q_{\psi '})$ corresponds to two  separate networks called target policy and target value neural networks, respectively, used for stabilizing the learning. At each iteration, they are gradually updated as the weighted sum between the current policy/value networks and the current target policy/value network, i.e., $\theta' \leftarrow (1-m_{target}) \theta' + m_{target} \theta$ and $\psi' \leftarrow (1-m_{target}) \psi' + m_{target} \psi$.
        
        \subsubsection{Distributional DDPG} 
        Another way is to approximate the distribution instead of only approximating the expected value of the return, as in \cite{DistributedDistributionalDDPG}. The following analogy is helpful here to motivate its interest. Instead of having a scheduler and its users, consider a teacher and its students. Even though the objective of the teacher is to increase the average ``knowledge'' of its students, using the distribution of the capacity/knowledge of the students allows for instance to decide whether to distribute his/her attention uniformly among students or to focus mostly on a fraction of them needing further support. 
        
        Algorithmically, it is impossible to represent the full space of probability distribution with a finite number of parameters, so the value neural network $\mathcal{Z}_{\psi}^{\pi_\theta}: \mathcal{S}\times\mathcal{A}\rightarrow \mathbbm{R}^{N_Q}$ is designed to approximate the actual $Z^{\pi_\theta}$ with a discrete representation. Among many variations \cite{CategoricalDQN,IQNdistributionalRL}, we choose the representation to be a uniform (discrete) probability distribution supported at $\{ (\mathcal{Z}_{\psi}^{\pi_\theta} )_i , i\in\{1,\cdots,N_Q\}\}$ where $(\mathcal{Z}_{\psi}^{\pi_\theta} )_i$ is the $i$-th element of the output. More rigorously, the distribution that the value neural network represents is 
        $\frac{1}{N_Q}\sum_{i=1}^{N_Q} \delta_{(\mathcal{Z}_{\psi}^{\pi_\theta} )_i}$, where $\delta_x$ is a Dirac delta function at $x$ \cite{QR-DQN}. Minimizing the 1-Wasserstein distance between this (approximated) distribution and the actual one of $Z^{\pi_\theta}$ can be achieved by minimizing the quantile regression loss
        \begin{equation}\label{eq:L1} 
        \mathcal{L}_1(\psi) = \sum_{i=1}^{N_Q}\mathbbm{E}_{ s_{t_0}\sim\mathrm{p}_{t_0},s\sim\rho^{\pi_\theta}_{s_{t_0}},z\sim \mathcal{T}^{\pi_{\theta'}}\mathcal{Z}_{\psi'}^{\pi_{\theta'}} (s,a) }[f_i(z-(\mathcal{Z}_{\psi}^{\pi_\theta} )_i) )]
        \end{equation}
        where $f_i(x){=}x(\frac{2i-1}{2N_Q}{-}\mathds{1}_{\{x<0\}})$, $\mathcal{T}^{\pi} Z^\pi (s,a) {\stackrel{D}{=}} R(s,a) {+} \gamma Z^\pi (s',\pi (s)), s'{\sim} P(s,a)$  is the distributional Bellman operator and $\mathcal{Z}_{\psi'}^{\pi_{\theta'}}$ is the target policy network (defined as before). 
        
\begin{figure*}
            \centering
            \begin{subfigure}[b]{0.31\textwidth}
            \centering
            \includegraphics[width=\textwidth]{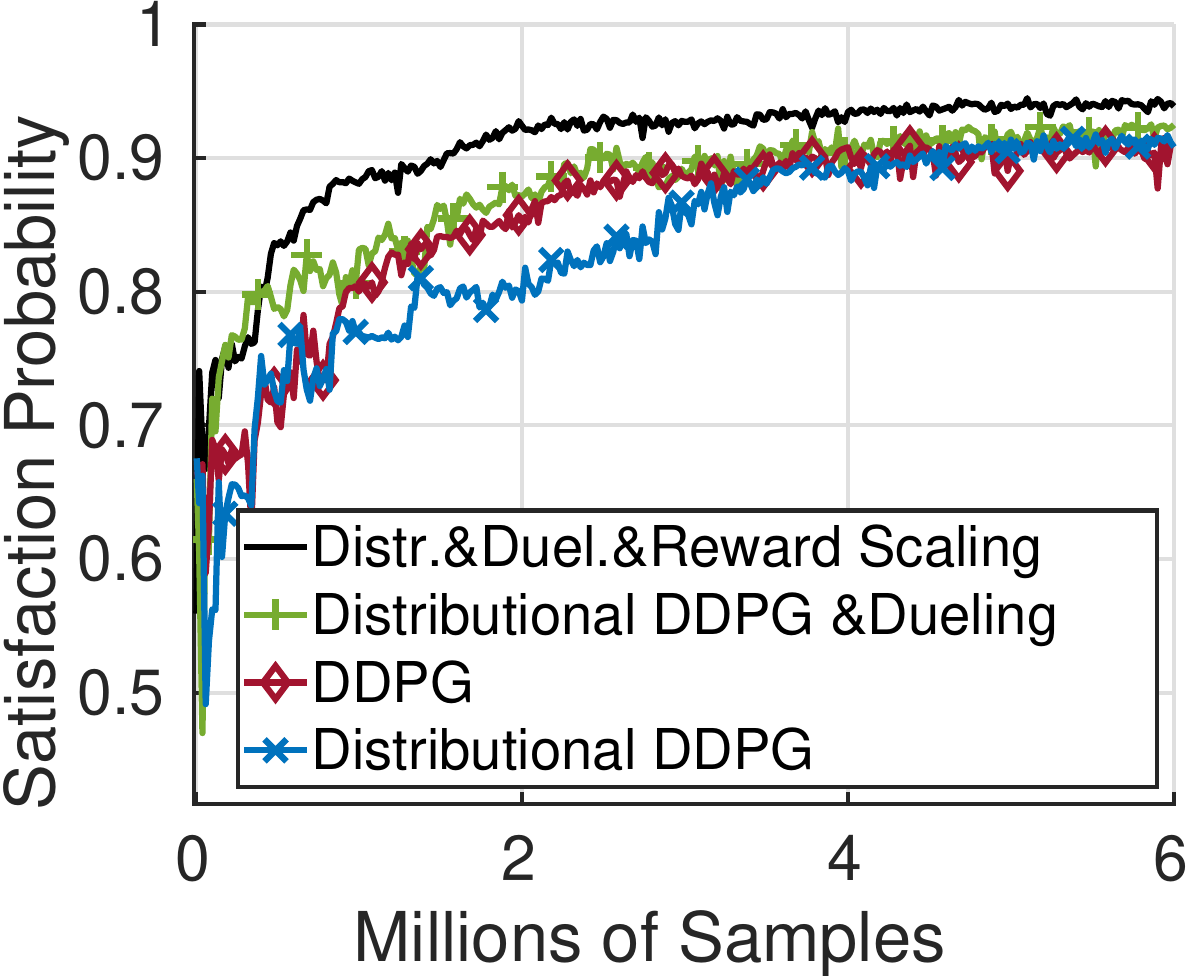}
            \caption{}
            \label{fig:Convergence Speed}
            \end{subfigure}
            \hfill
            \begin{subfigure}[b]{0.32\textwidth}
            \centering
            \includegraphics[width=\textwidth]{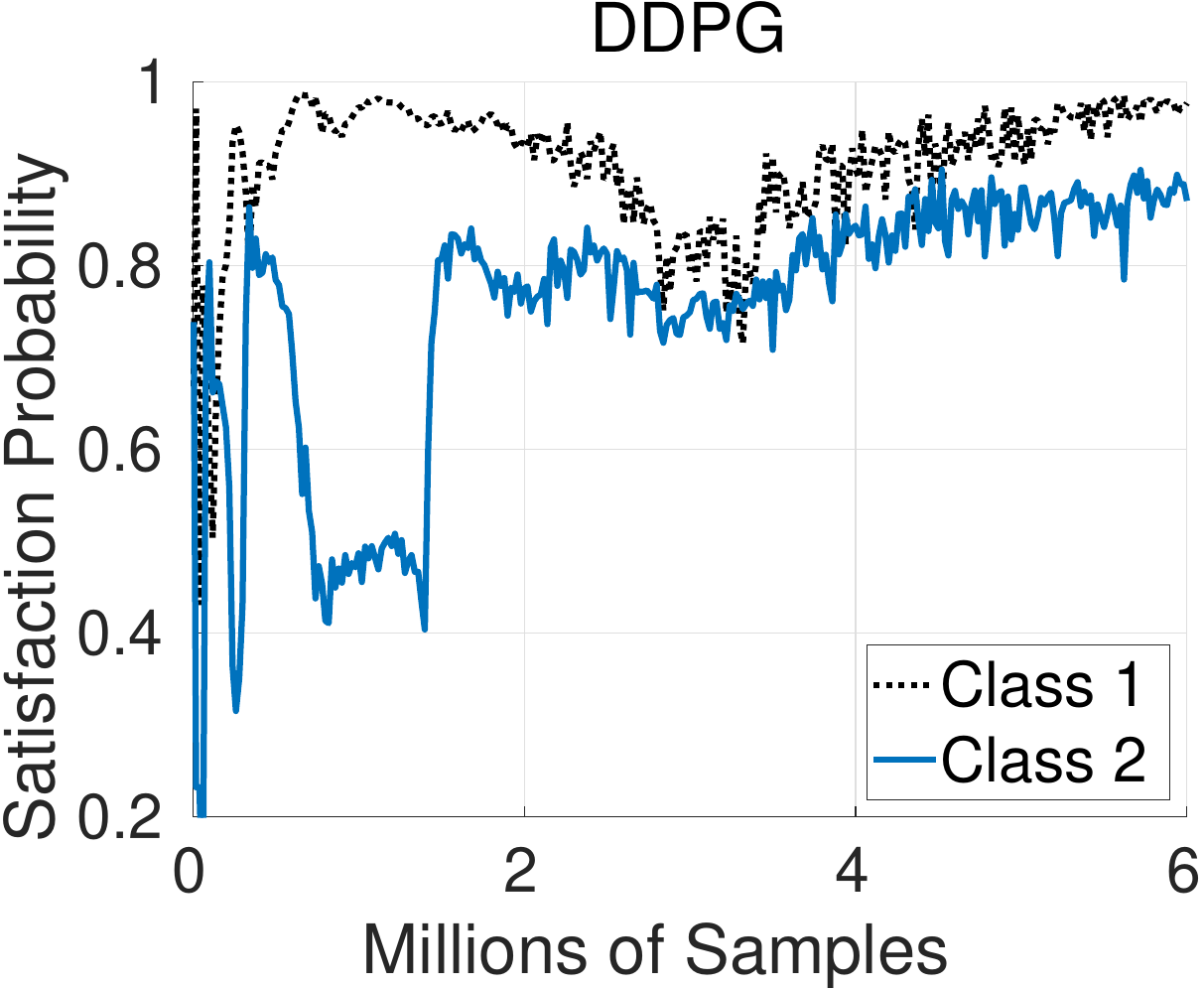}
            \caption{}
            \label{fig:L2UserClass1vsClass2}
            \end{subfigure}
            \hfill
            \begin{subfigure}[b]{0.33\textwidth}
            \centering
            \includegraphics[width=\textwidth]{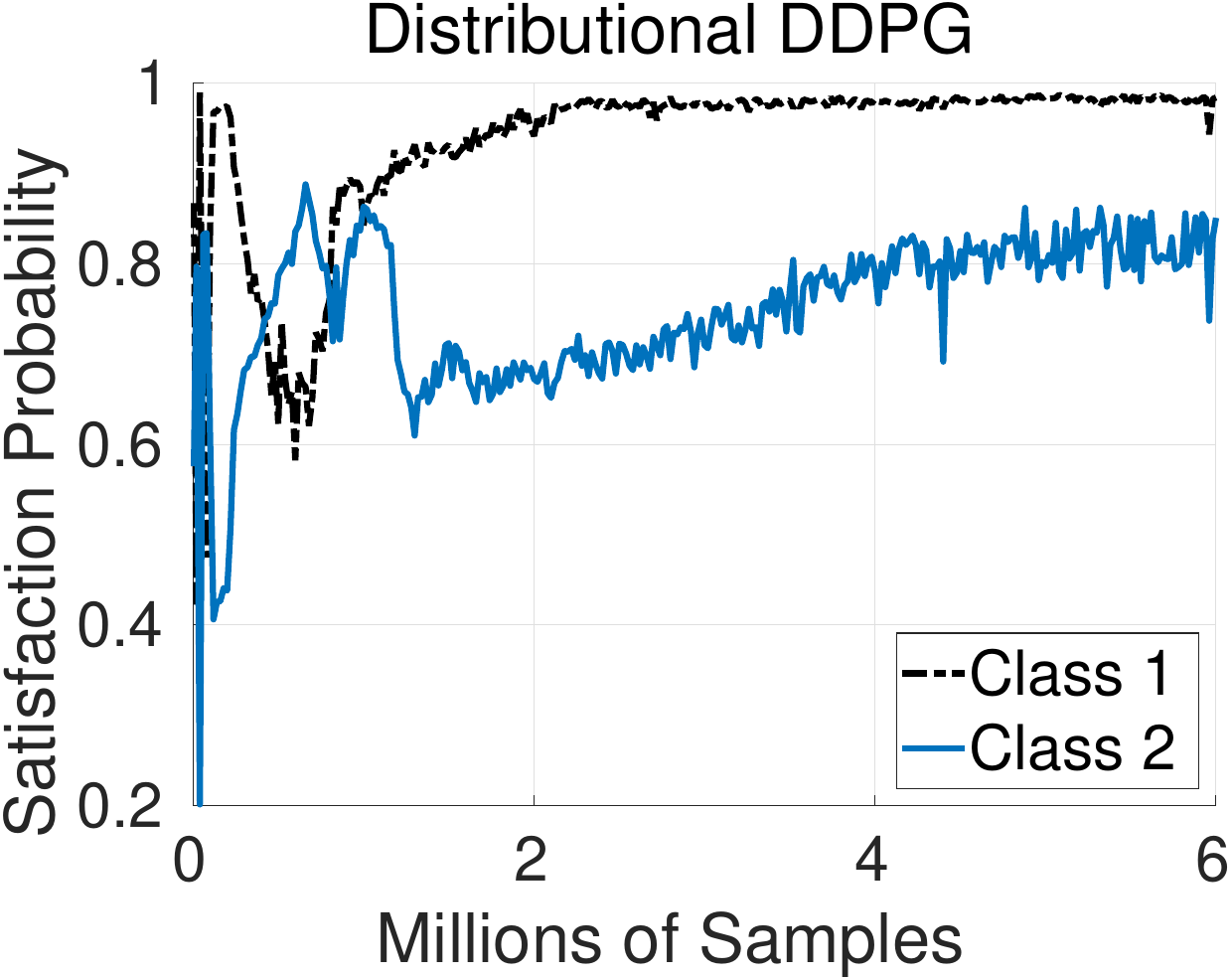}
            \caption{}
            \label{fig:L1UserClass1vsClass2}
            \end{subfigure}
            \hfill
            \caption{Comparison between distributional and standard (non-distributional) DDPG RL. We conducted five experiments with different seeds as in Figure \ref{fig:WithoutDeepSetsORNormal} with the same traffic model. In the first figure, we depict the average over those five experiments; in the other figures, we consider one specific experiment in an attempt to show the inherent ability of distributional DDPG in dealing with heterogeneous traffic.}
            \label{fig:ComparisonDistr}
        \end{figure*}

        Notice that even though we approximate the distribution of $Z^{\pi_\theta}(s,a)$, what is actually needed for improving the policy is only its expected value, approximated as $Q^{\pi_\theta}(s,a)\approx \frac{1}{N_Q}\sum_{i=1}^{N_Q} (\mathcal{Z}_{\psi}^{\pi_\theta} )_i$. Therefore it is natural to wonder if it indeed helps using $\mathcal{Z}_{\psi}^{\pi_\theta}$ instead of  directly approximating the needed expected value (confirming the intuition in the teacher-student analogy). In Figure \ref{fig:ComparisonDistr} we provide numerical support for distributional DDPG. Comparing Figures \ref{fig:L2UserClass1vsClass2} and \ref{fig:L1UserClass1vsClass2}, we show the benefits of using distributional DDPG. The distributional DDPG approach detects faster the existence of two different service classes with heterogeneous requirements, thus gradually improving the satisfaction rate for both of them. On the other hand, trying only to learn the expected value leads to a training where the performance for one class is improved at the expense of the other. Nonetheless, when aggregating the rewards coming from both classes, we observe in Figure \ref{fig:Convergence Speed} faster convergence of DDPG than the distributional DDPG even though - when converged - the latter exhibits slightly better performance. Introducing a trick (explained later in the ``dueling" paragraph), the distributional DDPG approach can be enhanced and outperforms DDPG.

\subsubsection{Distributional DDPG \& Dueling}  
To facilitate the approximation of the distribution $Z^{\pi_\theta}(s_t,a_t)$, we propose to split it into two parts: one that estimates the mean $\mathcal{Z}_{\psi}^{\pi_\theta, Mean}$ and one that estimates the shape of the distribution $\mathcal{Z}_{\psi}^{\pi_\theta,Shape}$. For that, we use a \textit{dueling} architecture \cite{DuelingDRL} (shown in Figure \ref{fig: NetworkArchitecture}). The output becomes $(\mathcal{Z}_{\psi}^{\pi_\theta})_i {=} \mathcal{Z}_{\psi}^{\pi_\theta, Mean} {+} (\mathcal{Z}_{\psi}^{\pi_\theta, Shape})_i {-} \frac{1}{N_Q}\sum_{i=1}^{N_Q}  (\mathcal{Z}_{\psi}^{\pi_\theta,Shape})_i$, $\forall i\in \{1,\cdots,N_q\}$; this effectively pushes $\mathcal{Z}_{\psi}^{\pi_\theta, Mean}$ \textit{to approximate} $Q^{\pi_\theta}$ used for training the policy. To ensure the decomposition of the distribution into shape and mean, we add a loss term $\mathcal{L}_{shape}=(\frac{1}{N_Q}\sum_i (\mathcal{Z}_{\psi}^{\pi_\theta,Shape})_i)^2$, centering  $\mathcal{Z}_{\psi}^{\pi_\theta,Shape}$ around zero. The total loss function is
        \begin{equation}\label{eq:dd}
            \mathcal{L}_{1+duel}(\psi) = \mathcal{L}_1(\psi) + \mathcal{L}_{shape}(\psi). 
        \end{equation}
            
\begin{figure*}[ht!]
            \centering
            \includegraphics[width=\textwidth]{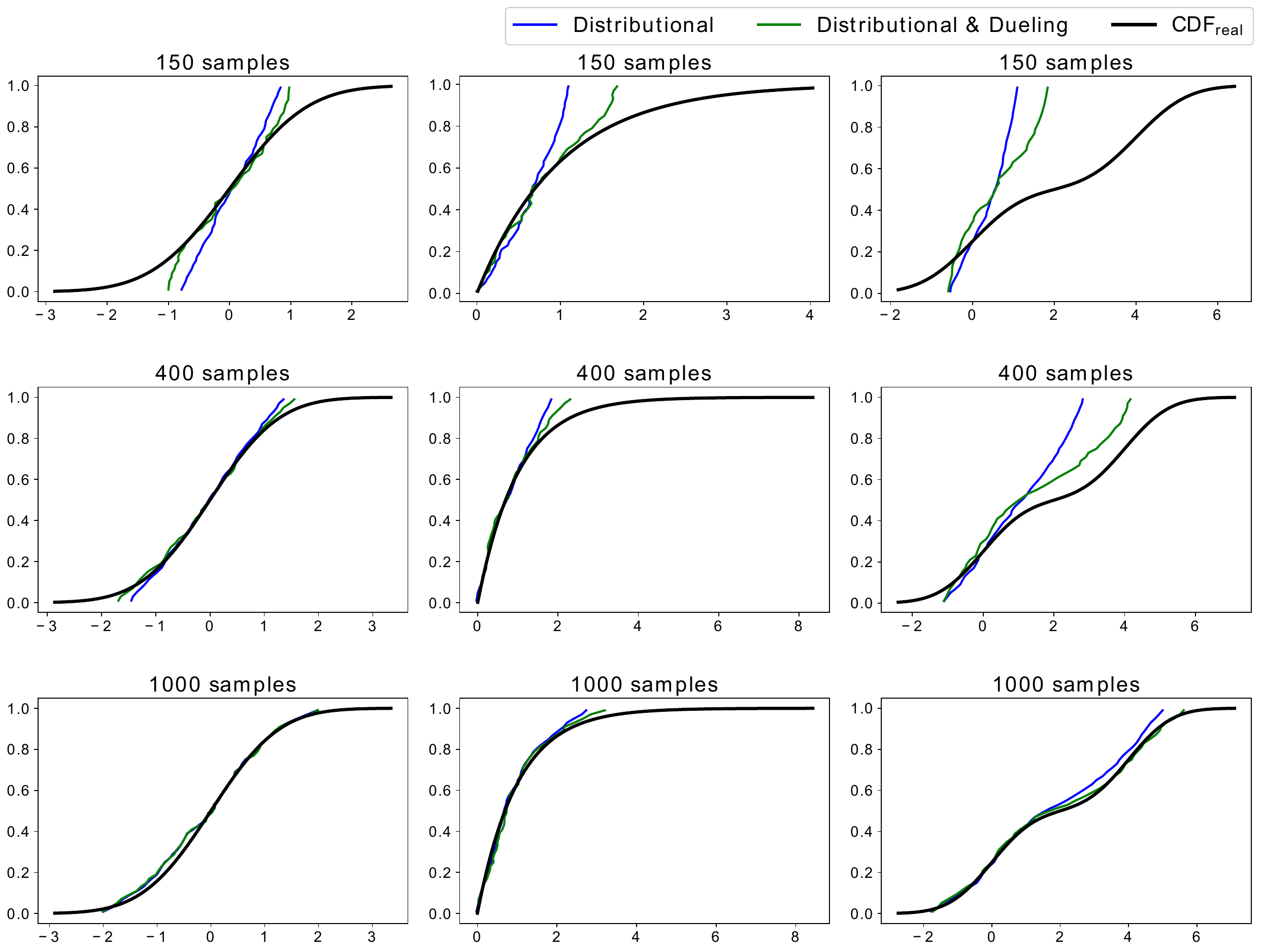}
            \caption{Estimation of a cumulative distribution function with and without the dueling trick.}
            \label{fig:Adam_dueling}
        \end{figure*}

To better understand the role and the performance of using the dueling architecture to approximate the (return) distribution, we have implemented a simple experiment, whose results are shown in Figure \ref{fig:Adam_dueling}. 
We set a random variable $Z$ with known cumulative distribution function (cdf) $\mathrm{CDF}_{real}$ from which we draw samples. The objective is to test the distributional and the combination of distributional plus dueling  approach on how fast using samples from $Z$ they correctly estimate the $\mathrm{CDF}_{real}$. For the first approach (termed Distributional) we use $N_Q$ parameters $\bm{\varphi}\in \mathbbm{R}^{N_Q}$ and aim to approximate the quantiles of $\mathrm{CDF}_{real}$ through minimizing the quantile regression loss (as in (\ref{eq:L1})): 
$\mathcal{L}_1(\bm{\varphi}) {=} \quad\sum_{i=1}^{N_Q}\mathbbm{E}_{ z\sim Z}[f_i(z{-}(\bm{\varphi})_i )]$.
 
On the other hand, we use the dueling architecture (termed Distributional \& Dueling) with parameters $\bm{\varphi}_{shape}\in \mathbbm{R}^{N_Q}$ and $\varphi_{mean}\in \mathbbm{R}$. We want with $\bm{\varphi}_{duel}:=[\bm{\varphi}_{shape},\varphi_{mean}]$ to approximate the quantiles of $\mathrm{CDF}_{real}$ by minimizing the loss  $\mathcal{L}_{1+duel}(\bm{\varphi}_{duel})$ as defined in (\ref{eq:dd}). In Figure \ref{fig:Adam_dueling}, each column corresponds to a different cdf $\mathrm{CDF}_{real}$:
        \begin{itemize}
            \item the first column corresponds to a normal distribution $\mathcal{N}(0,1)$,
            \item the second one to a Gamma distribution $\Gamma(1,1)$, and
            \item the last one to an equiprobable mixture of two normal distributions $\mathcal{N}(0,1)$ and $\mathcal{N}(4,1)$.
        \end{itemize}  
Each row corresponds to a different number of samples used to estimate $\mathrm{CDF}_{real}$. We depict the estimated cdf when using or not the dueling trick and compared them to the true one. We use $N_Q=50$ and the optimization algorithm is Adam with learning rate 0.01. We can see that using dueling leads to faster estimation of the true cdf in all cases.

\subsubsection{Scaling rewards}
A closer look on the range of the possible rewards reveals that they have a very large range of possible values, starting from 0 (no user satisfied) to $K$ (maximum number of users satisfied assuming all classes have  equal importance $\alpha_c = 1$). Therefore both its mean and variance may take big values. This is accentuated for the returns since it is the (discounted) sum of many of those rewards. Therefore, approximating the returns which take a large range of values is demanding. Standard technique to facilitate the approximation is ``scaling'' the rewards. The rewards are normalized in a way that the returns take values on a more easy to approximate range. 
Given a path that a fixed agent have taken, one can compute the returns per time slot across that path. Scaling the rewards pushes the mean of those returns to zero and the variance to one.

Specifically, the implementation  of scaling the rewards involves first estimating the discounted sum of rewards $z_t \leftarrow \gamma z_{t-1}  + r_t$, then the running statistic of its mean $z_t^{mean} \leftarrow m_{scale}z_{t-1}^{mean}+(1-m_{scale})z_t$ and of its mean of squares $z_t^{squares} \leftarrow m_{scale}z_{t-1}^{squares}+(1-m_{scale})z_t^2$. Finally the scaled reward equals to $\displaystyle\frac{r_t - z_t^{mean}}{\sqrt{z_t^{squares} - (z_t^{mean})^2} }$. The DRL algorithm is fed with those rewards whose discounted sum over time is the return that the policy network is trained to predict. We fix $m_{scale}= 10^{-4}$. In Figure \ref{fig:Convergence Speed}  it is shown that reward normalization clearly provides additional boost in the performance.        
\begin{figure}[htb]
            \centering
            \includegraphics[width=0.6\textwidth]{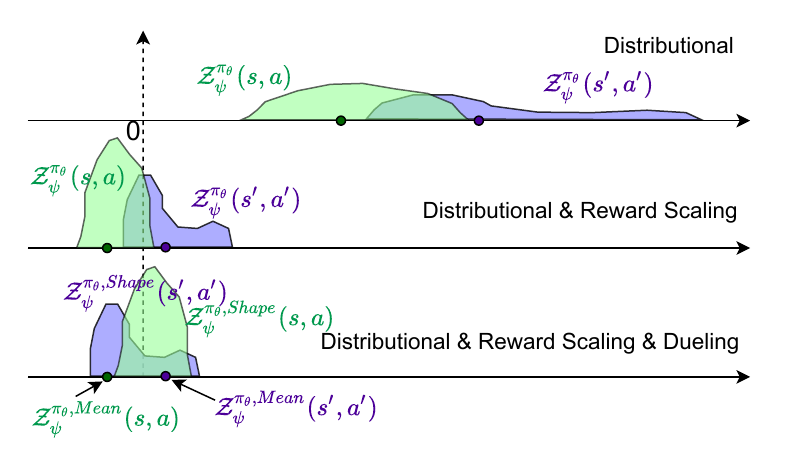}
            \caption{Effect of adding the dueling architecture in the Value Network and/or reward scaling to distributional DDPG RL.}
            \label{fig:RewardsNormalizationDuel}
\end{figure}
        
In Figure \ref{fig:RewardsNormalizationDuel}, we visualize what the value network tries to approximate. In the first row, by considering only distributional DDPG, from state $s$ and action $a$ the distribution of the returns $\mathcal{Z}^{\pi_\theta}_\psi(s,a)$ is approximated. From a different state $s'$ and action $a'$,  there will be other possible random paths that the agent with policy $\pi_\theta$ may take and the value network will try to approximate the distribution $\mathcal{Z}^{\pi_\theta}_\psi(s',a')$. The black dots depict the average of the two distributions, which are in fact the values that the value network of a simple DDPG would like to approximate and the policy network to maximize. In the second row, the use of reward scaling  shifts the distributions around zero and also shrink them. In the last row, the dueling trick is added so the value network has two outputs. One  branch of the dueling architecture approximates the value
$\mathcal{Z}^{\pi_\theta, Mean}_\psi(s,a) = \mathbbm{E}[\mathcal{Z}^{\pi_\theta}_\psi(s,a)]$, while the other the centered distribution $\mathcal{Z}^{\pi_\theta,Shape}_\psi(s,a) =\mathcal{Z}^{\pi_\theta}_\psi(s,a) -  \mathcal{Z}^{\pi_\theta, Mean}_\psi(s,a) $. 
        
\subsubsection{Deep Sets}
A final remark concerns the architecture, which, as discussed before, should be designed so as to preserve the permutation invariance. If we associate every user's characteristics with the resources given by the agent, i.e., the action corresponding to it, then permuting the users and accordingly the respective resource allocation should not influence the assessment of the success of the agent. To build such an architecture, we adopt the same architecture as in our Policy Network, capitalizing on ideas from DeepSets \cite{DeepSets}.
        
\begin{figure}[htb]
            \centering
            \includegraphics[width=1.0\textwidth]{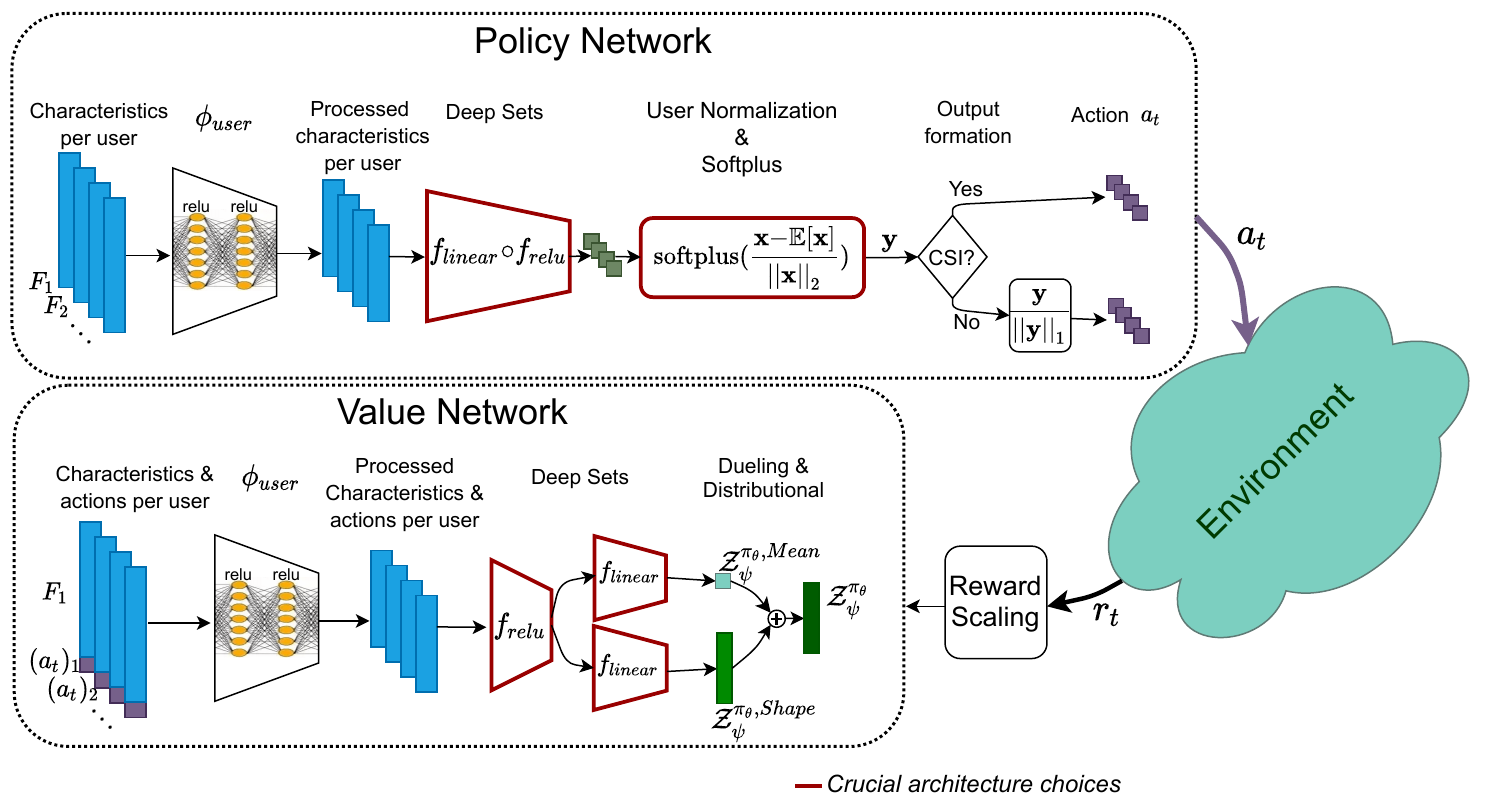}
            \caption{The proposed RL network architecture}
            \label{fig: NetworkArchitecture}
\end{figure}

The different steps of our algorithm are shown in Figure \ref{fig: NetworkArchitecture}.


\section{Baseline algorithms}\label{sec:baseline}
In this section, we present baseline scheduling algorithms, which are built upon conventional optimization techniques but are adapted to our specific problem. These algorithms are used for performance comparison in order to show the gains of our proposed DRL architecture.  
    
\subsection{Full CSI case}
At time $t_c~(t_c\geq  t_0)$, for user $u_0$, which arrived at time $t_0$, both channel $h_{u_0,t_c}$ and location $d_{u_0}$ are known. User $u_0$ is not satisfied at time $t$ if and only if the allocated bandwidth $w_{u_0,t}$ is smaller than the threshold $w_{u_0,t}^{th} =\displaystyle \frac{D_{u_0}}{\mathcal{R}_{u_0,t}}.$
We first consider algorithms working with immediate horizon ($T=1$), where only the current time $t_c$ is considered ignoring the effects on future slots. In that case, it is possible that the scheduler prefers serving two users that just arrived in the system rather than a user with bad channel requiring more resources but being on the verge of its latency constraint expiration. The optimization problem can be easily rewritten as follows. The variables to optimize are $\{x_{u,t_c}\}_u$. The variable $x_{u,t_c}$ is equal to $1$ if user $u$ is served at time $t_c$ or $0$ otherwise. The cost in terms of bandwidth used is $w_{u,t_c}^{th}x_{u,t_c}$, since full CSI is assumed and the scheduler allocates exactly the minimum bandwidth required to successfully send the data to user $u$. Then, the contribution in the reward function is $\alpha_u x_{u,t_c}$. As a result, the optimization problem can be written as 
    \begin{eqnarray*}
        &\underset{x_{u,t_c}}{\max} \quad  \sum_{u \in U_{t_c}} \alpha_u x_{u,t_c}\\
        &\text{s.t.} \quad \sum_{u \in U_{t_c}} w_{u,t_c}^{th} x_{u,t_c} \leq W  \\
        &\qquad x_{u,t_c}\in \{ 0,1\} ,\quad \forall u\in U_{t_c}.
    \end{eqnarray*}
This problem bils down to the {\it knapsack} problems, which aim to maximize the total value by choosing a proper subset from a set of objects. Every object has its value but also its weight, thus preventing one from picking all objects since the total weight of the chosen subset should not exceed the knapsack capacity level. It is a well known $\mathcal{NP}$-complete problem with numerous efficient algorithms solving it. In this work, we use Google's OR-TOOLS library for solving it.
    
    A second baseline we compare with is the so-called {\it exponential rule} \cite{ExponentialRule}, which corresponds to a generalization of proportional fair scheduler taking into account the queue state and the latency constraint of each user. At each time slot $t$, users are ordered according to their index values and we start serving the ones with the highest rank until resources are finished. Let $v_{u,t}$ be the number of the time slots user $u$ remains unsatisfied and $l_{u,t}$ be the number of time slots the user is eager to wait (therefore $L_u=v_{u,t}+l_{u,t}$). Denote $\displaystyle \overline{\mathcal{R}}_{u,t}=\frac{1}{v_{u,t}+1}\sum_{\tau = t-v_{u,t}}^t \mathcal{R}_{u,\tau}$ the estimated mean past rate. This value is known by the server at time $t$ by keeping track of the history of channel gains. Then the index $J_u$ for user $u$ is given by 
    $$\displaystyle J_u = \gamma_{u,t} \mathcal{R}_{u,t} e ^{ \displaystyle   \frac{a_{u,t}v_{u,t} - \overline{a_tv_t} }{1+\sqrt{\overline{a_tv_t}}  }   }$$
    with $\overline{a_tv_t}= \frac{1}{|U_t|}\sum_u a_{u,t}v_{u,t}$, $\gamma_{u,t}=a_{u,t}/\overline{\mathcal{R}}_{u,t}$ and $a_{u,t} = -\log(\delta_u)/l_{u,t}$ with $\delta_u$ being the delay violation probability.

    Lastly, we focus on algorithms that explicitly take into account the effects of an action on the future of finite horizon ($T>1$). For sake of simplicity, we assume that for the time interval $t\in[t_c,t_c+T-1]$ for all the current users and also for the ones that will appear within that interval, the channel realizations during this time interval are known beforehand (i.e., when the algorithm is executed at time $t_c$). Therefore, this baseline becomes an {\it oracle} since it knows the future channel realizations of users and can choose the best moment to serve them. Evidently, this method provides an \textit{upper bound} on the performance. Specifically, if $U^T_{t_c}$ denotes the set of all current users plus the ones that will arrive in the time interval $[t_c,t_c+T-1]$, then for every user $u\in U^T_{t_c}$ this baseline knows $w^{th}_{u,t}$ which corresponds to the required bandwidth  in order to satisfy $u$ at time $t\in[t_c,t_c+T-1]$. The optimization problem is then cast as 
    \begin{eqnarray*}
        &\underset{x_{u,t}}{\max} \quad  \sum_{u\in U^{T}_{t_c}} \alpha_u \sum_{t=t_c}^{t_c{+}T{-}1} x_{u,t}\\
        &\text{s.t.} \quad \sum_{U_t} w_{u,t}^{th} x_{u,t} \leq W,\quad \forall t\in[t_c, t_c{+}T{-}1]  \\
        &\qquad  \sum_{t=t_c}^{t_c{+}T{-}1} x_{u,t}\leq 1,\quad  \forall u\in U^{T}_{t_c}\\
        &\qquad x_{u,t}\in \{ 0,1\} ,\quad \forall t\in[t_c, t_c{+}T{-}1] \text{ and } \forall u\in U^T_{t_c}.
    \end{eqnarray*}
    This problem is an ILP and we use IBM CPLEX Optimization software, which employs the Branch and Cut algorithm~\cite{BranchAndCutIntro}, to solve it. 
    Notice that the above problem  cannot be mapped into a knapsack one (as for $T=1$), or even a  multiple knapsack problem because the weight of each user is time-varying due to channel variability and non-constant user set.

\subsection{Statistical CSI case}
Under statistical CSI, the BS knows the statistics of the system (channel, location, and traffic). In this section, we build a baseline to compare with the proposed DRL scheduler in the {\it no-CSI} case as mentioned in Remark \ref{rem:1}.
        
Let us first focus on the case of a single user $u_0$ arriving at time $t_0$. The current time is $t_c\in[t_0,t_0+L_{u_0}-1]$. We denote by ${\bm{w}}_{u_0,t}=(w_{u_0,t_0},$ $w_{u_0,t_0+1}, ... ,w_{u_0,t})$ the assigned bandwidth from time $t_0$ (beginning of transmission for user $u_0$). Additionally, let $A_{u_0,t}$ be a binary random variable, where if $A_{u_0,t}=1$, then $u_0$ is still unsatisfied at the end of time slot $t$ (\textit{after} receiving ${\bm{w}}_{u_0,t}$ resources) and $A_{u_0,t}=0$ otherwise. Given that at the beginning of time $t$, user $u_0$ still remains unsatisfied and that we know $w_{u_0,t}$ is scheduled at time $t$, we define $\Phi({\bm{w}}_{u_0,t}; d_{u_0})$ to be the probability that $w_{u_0,t}$ is not sufficient to satisfy the user's request for known location $d_{u_0}$ and unknown channel realization $h_{u_0,t}$, i.e., 
    \begin{equation}\label{eq:UserFailProbCurrent}
        \Phi({\bm{w}}_{u_0,t};\mathrm{d}_{u_0})=
           \left\{ \begin{array}{ll}
            \mathbbm{P}(A_{u_0,t}=1|{\bm{w}}_{u_0,t-1},\mathrm{d}_{u_0},A_{u_0,t-1}{=}1), & t>t_0\\
            \mathbbm{P}(A_{u_0,t}=1|\mathrm{d}_{u_0}), & t=t_c=t_0.
            \end{array}\right.
    \end{equation}
The average contribution of user $u_0$ to the gain \eqref{eq:GainFunc} on the time interval $[t_c,t]$ is given by the following equation, derived by applying the chain rule on conditional probability:
    \begin{equation}\label{eq:UserExpGain}
          \mathfrak{g}_{u_0}^{[t_c,t]}:=\mathfrak{g}(w_{u_0,t_c},...,w_{u_0,t};\mathrm{d}_{u_0}) = \left\{   \begin{array}{ll}
            0, \text{if } t_c>t_0 \text{ and }  A_{u_0,t_c-1}=0 \\
           {\displaystyle \alpha_{u_0}   \big( 1{-}\prod_{j=t_c}^t \Phi({\bm{w}}_{u_0,j};\mathrm{d}_{u_0}) \big) },  \text{ else.}
         \end{array}\right.
    \end{equation}
        
We consider now the average contribution on the gain \eqref{eq:GainFunc} for subsequent users after user $u_0$. The next user (if any) appears at time $t_1=t_0+L_{u_0}$, the second next at time $t_2=t_1+L_{u_1}$, and so on. In other words, we consider the users, denoted $u_1, u_2, \ldots$, which appear at time $t_1=t_0+L_{u_0}, t_2=t_1+L_{u_1}, \ldots$, respectively. Users belong to classes $c_1, c_2, \ldots$, with probabilities $p_{c_1}, p_{c_2}, \ldots$, respectively.  Since the locations of those future users are unknown, we need to average \eqref{eq:UserFailProbCurrent} and \eqref{eq:UserExpGain}  over their possible locations in order to obtain their contribution on the gain function \eqref{eq:GainFunc}. 
    So for $i\geq 1$ if ${\bm{w}}_{u_i,t}=(w_{u_i,t_i},$ $w_{u_i,t_i+1}, ... ,w_{u_i,t})$, we have 
    \begin{eqnarray}
          \mathfrak{g}_{u_i}^{[t_i,t]}{=}\mathfrak{g}(w_{u_i,t_i},...,w_{u_i,t}) =  
           {\displaystyle \alpha_{u_i}   \big( 1{-}\prod_{i=t_c}^t \Phi({\bm{w}}_{u_i,i}) \big) }
          \label{eq:UserExpGainFuture}
        \end{eqnarray}
        where the contribution looking at time $t$ with $t<t_i+L_{u_i}$ starts at time $t_i$ for user $u_i$ and where  
        \begin{equation}\label{eq:UserFailProbCurrentFuture}
            \Phi({\bm{w}}_{u_i,t})=\left\{
            \begin{array}{ll}
            \mathbbm{P}(A_{u_i,t}=1|{\bm{w}}_{u_i,t-1}, A_{u_i,t-1}{=}1),& \quad t>t_i\\
            \mathbbm{P}(A_{u_i,t}=1), & t=t_i.
            \end{array}\right.
        \end{equation}
    Closed-form expressions for  Eqs. \eqref{eq:UserFailProbCurrent}  and \eqref{eq:UserFailProbCurrentFuture} are provided in Appendix \ref{anx}.
    
For notational convenience, to include the case where no new user is generated in a time slot, we introduce the ``null'' class of users, which contains users serving as dummies. They appear with probability $p_{null}$, are active for one slot ($L_{null}=1$) and have zero contribution $\mathfrak{g}_{u}^{[t_i,t_{i+1}]}=0$ with $t_{i+1} {=} t_i{+}L_{null}$.     
    Hence, the average value of the gain function for the sequence of users $u_0, u_1,...$ (so when there is one user at most per time slot, i.e., $K=1$) starting at the current time $t_c$ is
    \begin{eqnarray}
      &\mathcal{G}(w_{u_0,t_c},..., w_{u_0,t_1-1},w_{u_1,t_1},...) =\mathfrak{g}_{u_0}^{[t_c,t_1{-}1]}(.;d_{u_0})\nonumber \\
      & + \displaystyle{ \sum_{c_1\in\mathcal{C}\cup \mathrm{null}} }\bigg(p_{c_1}\cdot\mathfrak{g}_{u_1}^{[t_1,t_2{-}1]}+ \displaystyle{\sum_{c_2\in\mathcal{C}\cup \mathrm{null}}}\Big(p_{c_2}\cdot\mathfrak{g}_{u_2}^{[t_2,t_3{-}1]} +  \displaystyle{\sum_{c_3\in \mathcal{C}\cup \mathrm{null}}}(...)\Big)\bigg).\qquad
      \label{eq:OneServer}
    \end{eqnarray}
From \eqref{eq:OneServer}, we observe a tree structure\footnote{This can be exploited for computing it recursively.} that when a user vanishes, there is a summation over all possible classes the new user may belong to. Therefore, a number of branches equal to the number of possible classes (equal to $|\mathcal{C}|$) are created whenever a new future user is taken into account. To harness this scalability issue, we prune the tree by considering only $T$ future time slots and work with finite horizon $[t_c,t_c+T-1]$. 

    The general case with multiple users served simultaneously ($K > 1$) can easily be considered by just computing $K$ ``parallel trees". With a slight abuse of notation, we consider that the first subscript of the variables $w$ refers now to the index of the tree (and implicitly to a specific user). As a consequence, the variables for the scheduled bandwidth resources over an horizon of length $T$ can be put into the following matrix form
    \begin{equation*}
        \mathbf{W}_{t_c}=
        \begin{bmatrix}
        w_{1,t_c} & w_{1,t_c+1} & \cdots & w_{1,t_c+T-1} \\
        w_{2,t_c} & w_{2,t_c+1} & \cdots & w_{2,t_c+T-1} \\
        \vdots  & \vdots  & \ddots & \vdots  \\
        w_{K,t_c} & w_{K,t_c+1} & \cdots & w_{K,t_c+T-1} 
        \end{bmatrix}
    \end{equation*}
    and the average gain for these resources takes the following form:
    \begin{align}
        G(\mathbf{W}_{t_c}) = \sum_{k=1}^K \mathcal{G}(w_{k,t_c}, w_{k,t_c+1}, \cdots, w_{k,t_c+T-1}).\label{eq:objFunc}
    \end{align}
    Finally, we arrive at our optimization problem at current time $t_c$:
    \begin{align}
        &\underset{\mathbf{W}_{t_c}\in\mathbbm{R}^{K\times T}_{\geq 0}}{\max} \quad  G(\mathbf{W}_{t_c})\\
        & \text{s.t.} \quad \sum_{k=1}^K w_{k,t} \leq W,\quad \forall t\in\lbrace t_c, \ldots, t_c{+}T{-}1\rbrace. \label{eq:constraintG}
    \end{align}
    
    It can easily be shown that the objective function $G(\cdot)$ is non-concave with multiple local optima. The constraints given by \eqref{eq:constraintG} describe a compact and convex domain set, which allows applying the Frank-Wolfe algorithm (FW) \cite{FrankWolfe} that guarantees reaching to a local optimum. The convergence of the FW method is {\it sublinear}; however, computing the objective function \eqref{eq:objFunc} and its partial derivatives grows {\it exponentially} with $T$, thus leading to slow and cumbersome method in practice. Therefore, in each time slot $t_c$, we use FW to get a local optimum solution $\mathbf{W}_{t_c}^\star$ from which we retrieve the first column $[w_{1,t_c}^\star,\cdots,w_{K,t_c}^\star]^\intercal$ corresponding to the bandwidth allocation that will be applied at the current time step $t_c$.

\section{Experimental Results}\label{sec:experiments}
  
\subsection{Synthetic Data}
We consider the distance-dependent pathloss model $120.9+37.6\log_{10}\mathrm{d}$ (in dB) \cite{LTEstandard}, which corresponds to a constant loss component $C_{pl}=10^{-12.09}$ and pathloss exponent $n_{pl}=3.76$. The noise spectral density is $\sigma_N^2 = -149$dBm$/$Hz. We consider that the distance between the base station and users ranges from 0.05 km to 1 km. The power per unit bandwidth is kept equal to $1 \mu$W$/$Hz.
      
    For the proposed DRL scheduler, we update the target policy and value networks with momentum $m_{target}=0.005$. We use replay buffer of capacity $5000$ samples. The batch size is set to $64$ and the learning rate is set to $0.001$. The discount factor is $\gamma=0.95$.  We use $N_Q=50$ quantiles to describe the distribution. The $\phi_{user}$ consists of two fully connected layers each with $10$ neurons. We have $P_{explore}=0.2$ and $\sigma_{explore} = 0.3$. The number of input and output dimensions  in both $f_{\mathrm{relu}}$ and $f_{\mathrm{linear}}$ is 10 (i.e., $H=H'=10$). We remark that the number of parameters is kept relatively low (around 1800), mainly due to the use of Deep Set. Increasing this further unavoidably results in overfitting due to the high stochasticity of the environment. Moreover, keeping the number of parameters low makes our solution fast and cost-effective (both in terms of energy and hardware). 
    
    We consider two scenarios for the traffic as described in Table \ref{tab:classesDescription}. 
    \begin{table}[htb]
    \caption{Classes description for two scenarios}\label{tab:classesDescription}
        \begin{subtable}[h]{1\columnwidth}
            \centering
            \caption{Users of equal importance}
            \label{tab:Equal Users}
            \begin{tabular}{c||c|c|c|c}
            $ $ & Data per user (Kbytes) & Latency (in time slots)& Imp. & Prob.\\
            \hline
            Class 1    &  8 & 2 & 1 & 0.3\\
            Class 2    &  64 & 10 & 1 & 0.2
            \end{tabular}
        \end{subtable}
        \bigskip
        \begin{subtable}[h]{1\columnwidth}
            \centering
            \caption{Prioritized and normal users }\label{tab:Inequal Users}
            \begin{tabular}{c||c|c|c|c}
            $ $ & Data per user (Kbytes) & Latency (in time slots)& Imp. & Prob.\\
            \hline
            Class 1    &  8 & 2 & 1 & 0.15\\
            Class 1+    &  8 & 2 & 2 & 0.05\\
            Class 2    &  64 & 10 & 1 & 0.3\\
            Class 2+    &  64 & 10 & 2 & 0.05
            \end{tabular}
        \end{subtable}
    \end{table}
    
The first scenario consists of two classes, one with users requesting a small amount of data but within a stringent latency constraint (of just two time slots) and one delay-tolerant class requesting a large amount of data. All classes have the same importance as seen from the Imp. column. In the second scenario, classes do not have the same importance.
Note that the Prob. column describes the probability $p_c$ with which a user of that class appears in the system at a given time slot (they do not sum up to one signifying that it is possible that no user appears during some time slots).
    
\begin{figure*}[ph!]
         \centering
         \begin{subfigure}[b]{0.48\textwidth}
             \centering
             \includegraphics[width=1.05\textwidth]{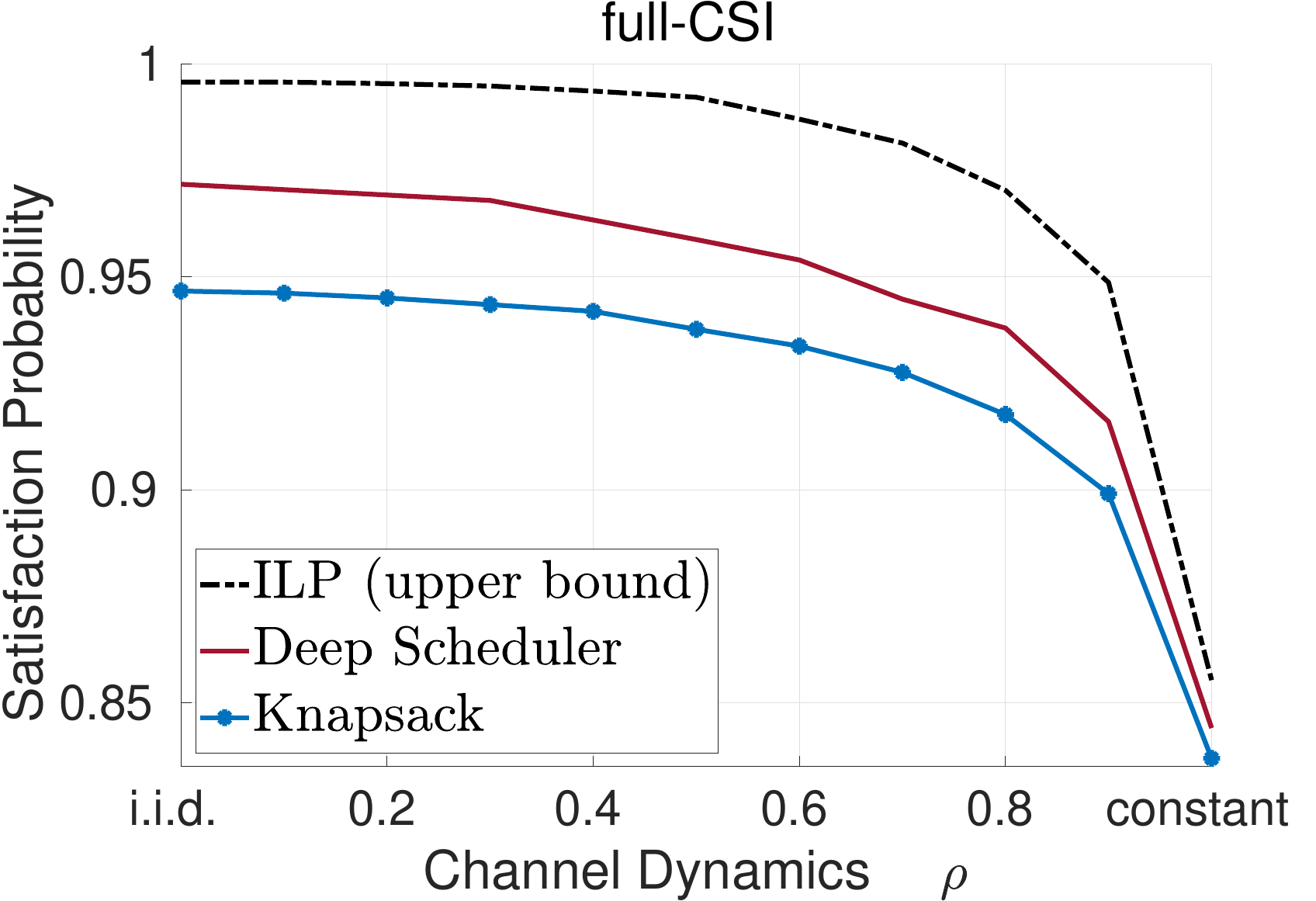}
             \caption{}
             \label{fig:fullCSI_vsRho}
         \end{subfigure}
         \begin{subfigure}[b]{0.48\textwidth}
             \label{subfig:Full_vs_bw}
             \centering
             \includegraphics[width=0.97\textwidth]{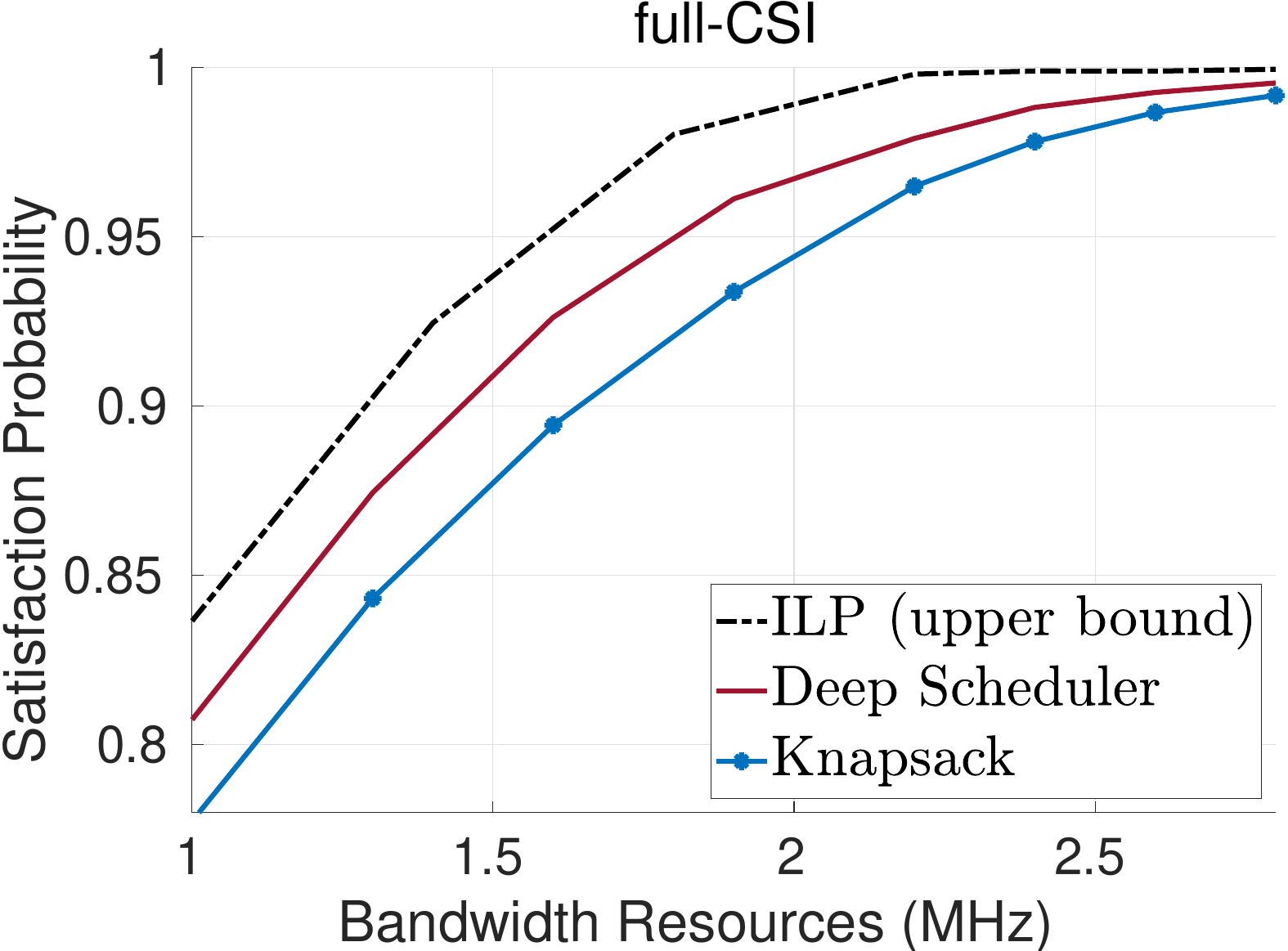}
             \caption{}
             \label{fig:fullCSI_vsBw}
         \end{subfigure}
         \hfill
         \begin{subfigure}[b]{0.48\textwidth}
             \centering
             \includegraphics[width=1.05\textwidth]{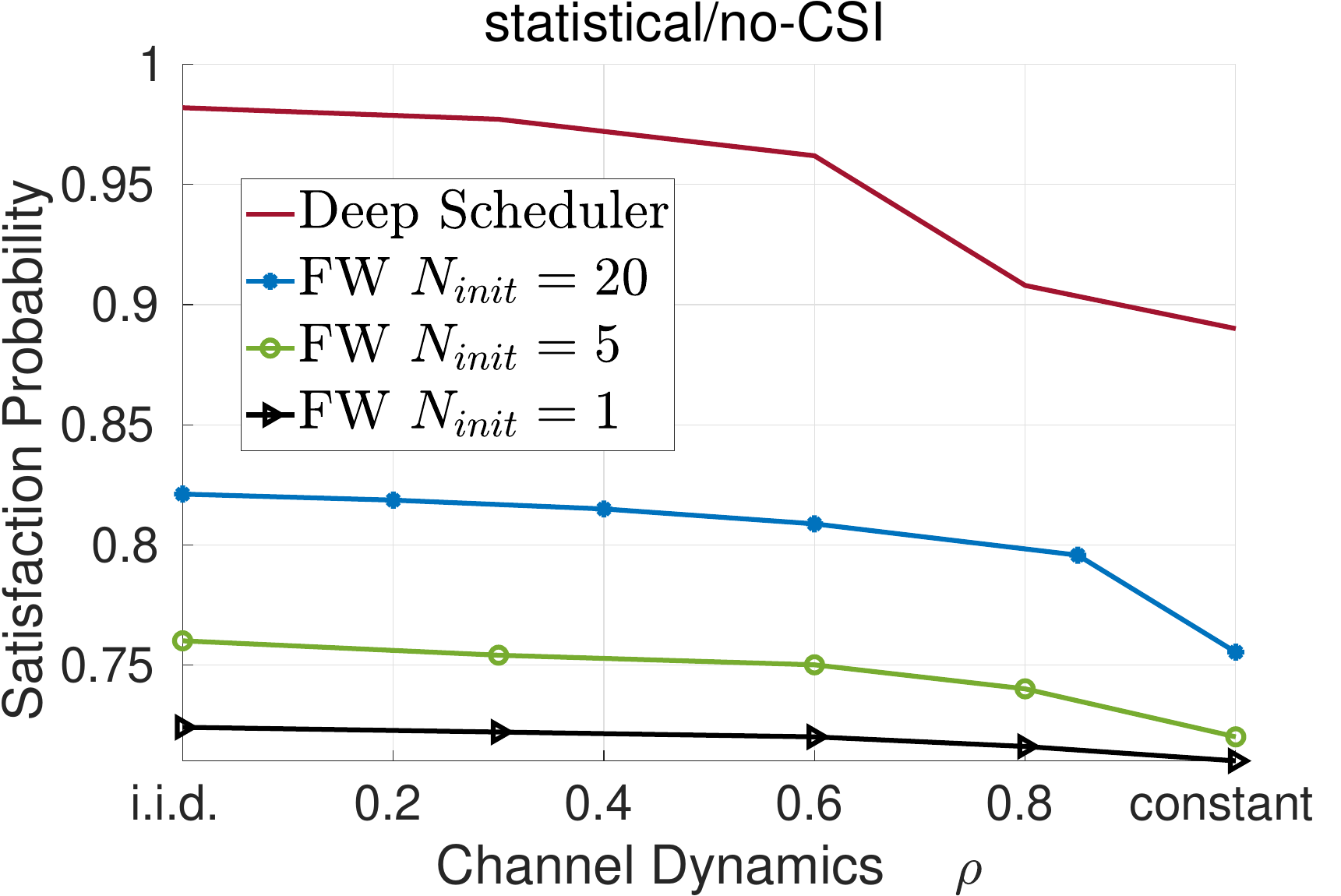}
             \caption{}
             \label{fig:StatCSI_vsRho}
         \end{subfigure}
         \centering
         \begin{subfigure}[b]{0.48\textwidth}
             \centering
             \includegraphics[width=0.97\textwidth]{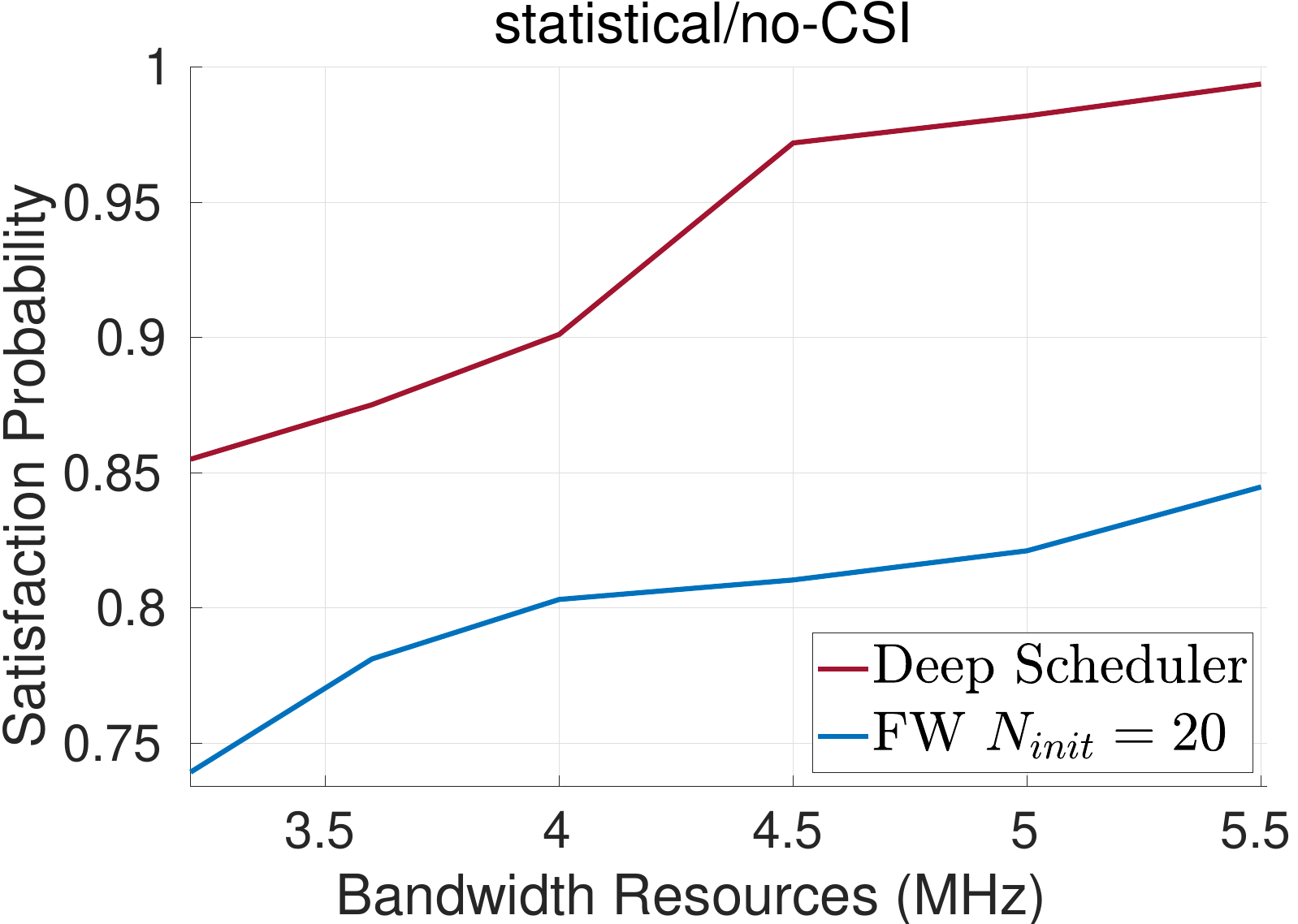}
             \caption{}
             \label{fig:StatCSI_vsBw}
         \end{subfigure}
         \hfill
         \begin{subfigure}[b]{0.48\textwidth}
             \centering
             \includegraphics[width=1.05\textwidth]{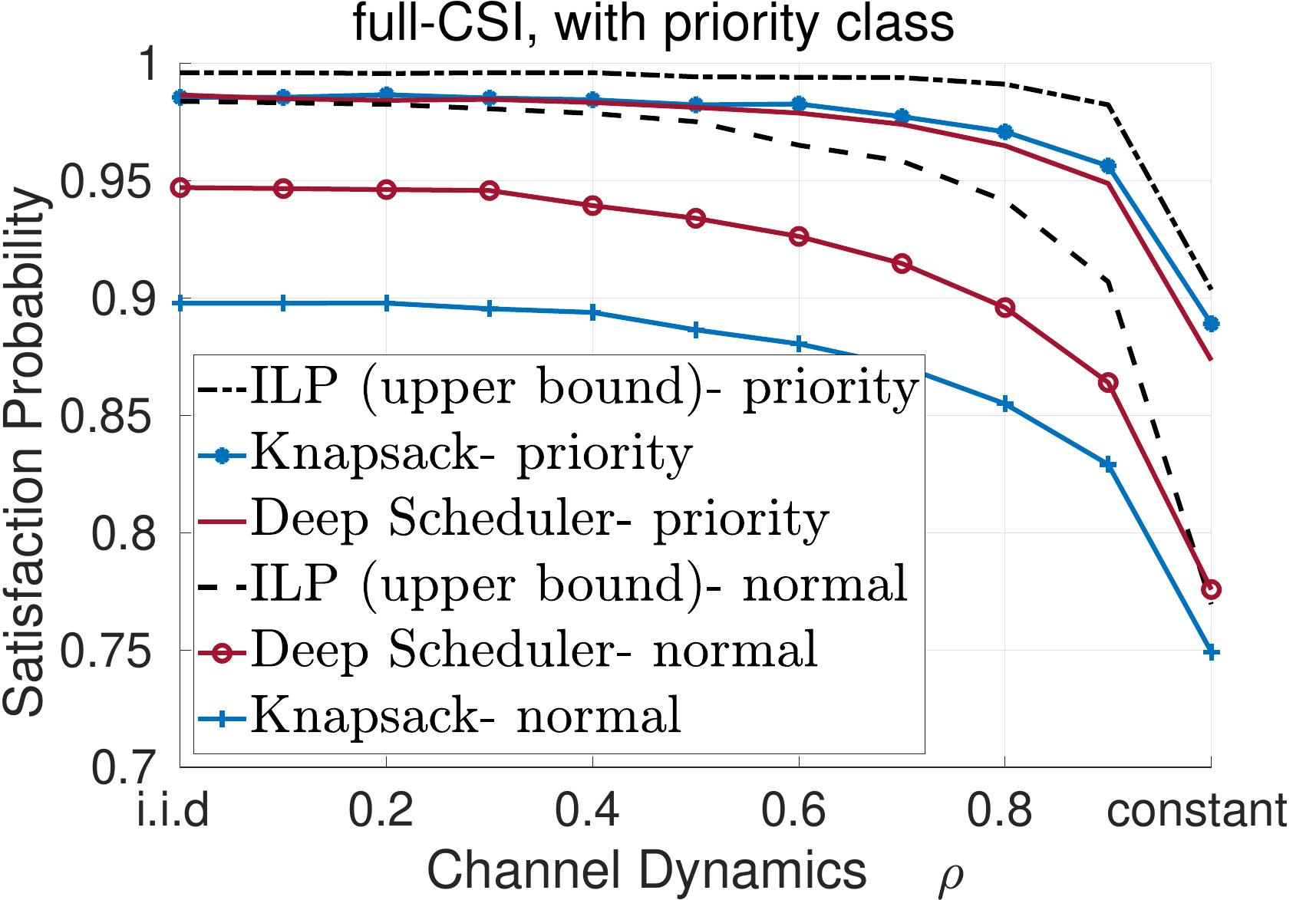}
             \caption{}
            \label{fig:fullCSI_vsRho_privileged}
         \end{subfigure}
         \begin{subfigure}[b]{0.48\textwidth}
             \centering
             \includegraphics[width=0.97\textwidth]{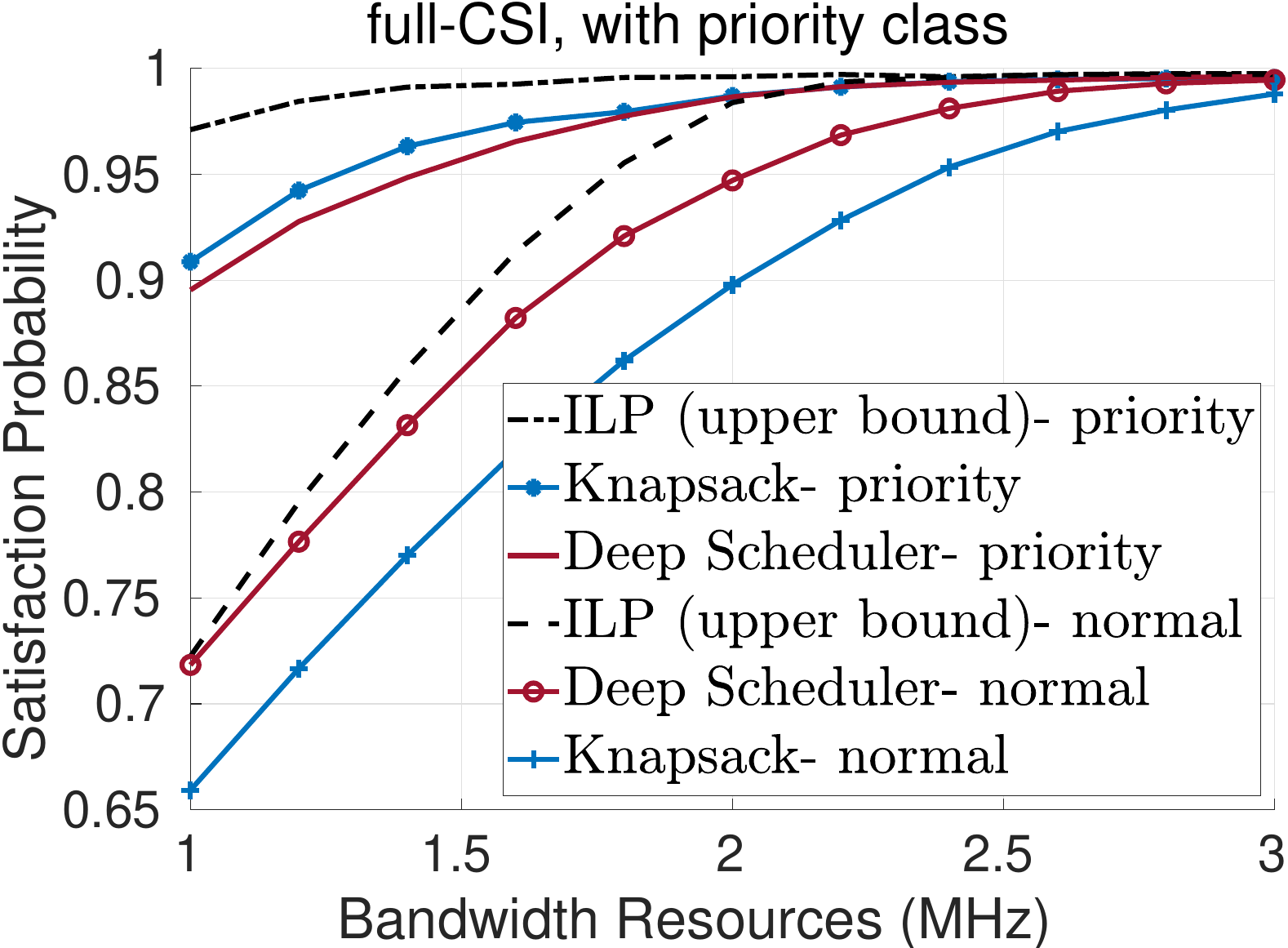}
             \caption{}
             \label{fig:fullCSI_vsBw_privileged}
         \end{subfigure}
         \hfill
         \caption{Satisfaction rate of the proposed DRL scheduler and the baseline algorithms versus $\rho$ (left column) and $W$ (right column). The first and last rows correspond to the case of Table \ref{tab:Equal Users} and the row in the middle to Table \ref{tab:Inequal Users}. Figures \ref{fig:fullCSI_vsRho}, \ref{fig:fullCSI_vsBw}, \ref{fig:fullCSI_vsRho_privileged} and \ref{fig:fullCSI_vsBw_privileged} refer to the full CSI case while the other to the statistical CSI/no CSI case.}\label{fig:perfs_synthetic}
    \end{figure*}

In Figure \ref{fig:perfs_synthetic}, we plot the satisfaction ratio per class priority (i.e., all users having the same priority take part to the computation of the same ratio and so depicted in the same curve) versus the channel correlation ($\rho$ in left column) and versus the total bandwidth ($W$ in right column) for both scenarios and different CSI knowledge. Figures \ref{fig:fullCSI_vsRho},\ref{fig:StatCSI_vsRho},\ref{fig:fullCSI_vsRho_privileged} are plotted for $W=2$ MHz, $W=5$ MHz and $W=2$ MHz, respectively, whereas Figures \ref{fig:fullCSI_vsBw}, \ref{fig:StatCSI_vsBw}, \ref{fig:fullCSI_vsBw_privileged} are all for $\rho = 0$. Figures \ref{fig:fullCSI_vsRho}, \ref{fig:fullCSI_vsBw}, \ref{fig:fullCSI_vsRho_privileged}, and \ref{fig:fullCSI_vsBw_privileged} are done with $K=100$ users. Figures \ref{fig:StatCSI_vsRho} and \ref{fig:StatCSI_vsBw} are done with $K=60$ users.    
    
Recall that the FW algorithm reaches to a suboptimal point and different initializations lead to different local optima. For that, at each time slot, we repeat the FW algorithm $N_{init}$ times with a different initialization at each time and we select the best suboptimal point. This method could lead to considerable performance improvement for $N_{init}$ increasing; however, due to computational complexity, we  stop  at $N_{init} = 20$. Moreover, as the number of users $K$ increases, so does the number of local optima and that of solutions with poor performance making it tougher for the FW to find a good optimal point without significantly increasing $N_{init}$. This is the main reason why our DRL Scheduler substantially outperforms Frank-Wolfe algorithm even at moderate values of users ($K=60$). Note that our DRL  Scheduler continues exhibiting very good performance even if $K$ is further increased.

The proposed DRL scheduler significantly outperforms the knapsack algorithm. For instance, at a level of $95\%$ of satisfaction probability, we may save about $13\%$ of bandwidth, which is followed by a $13\%$ power saving as the power per Hz is kept constant (see Figure \ref{fig:fullCSI_vsBw}). We also observe that our scheduler is quite close to the optimal policy, since ILP which uses an oracle constitutes an upper bound on the performance. In Figures \ref{fig:fullCSI_vsRho_privileged} and \ref{fig:fullCSI_vsRho_privileged}, there is a priority class of users, which always enjoys a higher satisfaction probability. Interestingly, the proposed DRL scheduler serves slightly worse the priority class than what the knapsack does. Nevertheless, since the priority counts for the $\frac{0.1}{0.55}\approx 18\%$ of the users and the rest $82$\% is much better served using our Scheduler, the latter exhibits overall better performance than the knapsack.

\subsection{Real Data}
\label{subsec:real data}
To assess the applicability of our algorithm in a realistic setup, we perform experiments on real data using publicly available traces based on real measurements over Long Term Evolution (LTE) 4G networks in a Belgium city \cite{BelgiumReal4GData,GithubPensieve}. Six different types of transportation (foot, bicycle, bus, tram, train, car) are used. The throughput and the GPS location of a mobile device continuously demanding data are recorded every second. Since the timescale of 1$s$ is much larger than the small-scale fading timescale represented here by the random variable $h$, the measurement that is provided corresponds to $M_i=\mathbbm{E}_h[W\log_2(1+\kappa |h|^2)]$ for every $i$-th second. The value of $\kappa$, which mainly depends on the user location, is assumed constant within 1$s$. As the measurements bandwidth $W$ is not reported in the dataset, we assume it to be $15$MHz, resulting in a mean signal-to-noise ratio (SNR) $\approx$ 6dB in an LTE compliant system. This allows us to retrieve $\kappa$ from measurement $M_i$. To compute the channel time variation $h$, the user speed is required in \eqref{eq:Small Scale Fading Markov} so as to obtain $\rho$. This is estimated using the trajectory of GPS coordinates given from the traces. A user entering the system belongs to a class according to Table \ref{tab:Table Classes for RealData}, with its type of transportation chosen randomly; we then sample $M_i$ and her location from the traces accordingly. Knowing in the previous and afterwards time slots the locations, we can compute the average speed and so the $\rho$. Finally, so far we assumed that the bandwidth can be split as small as desired (continuum); however, in practice, the bandwidth is split into $N_{bl}$ resource blocks and each user is assigned an integer multiple of those. In Table \ref{tab:Results, Real Data}, we increase $N_{bl}$ with the size of the resource block constant to $200$kHz, again confirming the performance gains from using a DRL based approach. For the Exponential rule, the value of $\delta_u$ is set to $\delta_u=\delta=10^{-2}$ \cite{ExponentialRule}. Nevertheless, since this value does not provide the best results for every $N_{bl}$ (resource blocks), we tune this parameter for each $N_{bl}$ in order to provide the highest possible performance.
     \begin{table}[htb]
     \centering
        \caption{Equal Classes description (Data rate per user in Kbps, Latency in msec)}
        \label{tab:Table Classes for RealData}
            \begin{tabular}{c||c|c|c|c}
                $ $ & Data (Kbits) & Latency (msec) & Imp. & Prob.\\
                \hline
                \hline
                Class 1    &  1 &  5 & 1& 0.2\\
                Class 2    &  5   & 25 & 1& 0.3
            \end{tabular}
    \end{table}
        
    \begin{table}[htb]
         \centering
                \caption{Sum Data Rate (in Mbps) / Probability of Satisfaction}\label{tab:Results, Real Data}
            \begin{tabular}{c||c|c|c|c|c}
                $N_{bl}\; (W)$             & 6 (3MHz)       & 15 (5MHz)     & 25 (10MHz)     & 50 (15MHz)     & 75 (20MHz)\\
                \hline\hline
                Knapsack     &  6.4 / 48.6\%       & 10.2 / 64.3\%      &  15.2 / 84.3\%       & 17.0 / 91.2\%        & 17.6 / 93.4\%\\
                Exp. Rule       &  5.9 / 44.0\%       & 8.8 / 57.3\%        &  14.0 / 80.4\%       & 17.2 / 92.7\%        & 18.1 / 95.8\%\\
                \textbf{Proposed DRL Scheduler}     &  \textbf{6.7 / 51.6}\%       & \textbf{10.6 / 67.6\%}       &  \textbf{15.5 / 86.9\% }      & \textbf{17.2 / 93.0\%}        & \textbf{18.3 / 96.2\%}\\
                ILP (Upper Bound)      &  9.0 / 62.9\%       & 14.6 / 81.8\%        &  18.6 / 98.5\%       & 18.9 / 98.7\%       & 19.0 / 98.9\%
            \end{tabular}
    \end{table}
    
In Table \ref{tab:Results, Real Data}, we see that the proposed DRL algorithm outperforms baseline algorithms with full CSI and using real data, both in terms of data rate and satisfaction probability. The gap from the upper bound is rather significant, but this is expected as the upper bound is optimistic assuming that the channel is known in advance.

\section{Conclusion}\label{sec:ccl}
The problem of scheduling and resource allocation of a time-varying set of users with heterogeneous traffic and QoS requirements was studied here. We leveraged deep reinforcement learning and proposed a deep deterministic policy gradient algorithm, which builds upon distributional reinforcement learning and deep sets. Our experiments on both synthetic and real data showed that the proposed scheduler can achieve significant performance gains as compared to state-of-the-art conventional combinatorial optimization methods in both full and no CSI scenarios.

\appendices
\section{Closed-form expressions for \eqref{eq:UserFailProbCurrent}  and \eqref{eq:UserFailProbCurrentFuture}}\label{anx}
  
\paragraph{i.i.d. fading~($\rho=0$)} 
This is the simplest case since there are no time dependencies on the fading, hence using \eqref{eq:FailureProb_u_t_given_dist} and \eqref{eq:FailureProb_u_t}, \eqref{eq:UserFailProbCurrent} and \eqref{eq:UserFailProbCurrentFuture} become 
\begin{align*}\displaystyle
\Phi({\bm{w}}_{u_0,t};\mathrm{d}_{u_0}) = P_{u_0}^{fail}(w_{u_0,t},P;\mathrm{d}_{u_0}) \textnormal{ and } \Phi({\bm{w}}_{u_i,t}) = P_{u_i}^{fail}(w_{u_i,t},P), \quad i\geq 1.
\end{align*}
We recall that users $u_i$ for $i\geq 1$ arrive after user $u_0$, therefore their locations are unknown and we need to average over them\footnote{It might not be easy to find the derivative of \eqref{eq:FailureProb_u_t}, which is required for first-order approximation in the Franck-Wolfe algorithm. This is done as follows
    \begin{align*}
    \frac{d P_{u}^{fail}}{dw} = \int_{\mathrm{d}_{min}}^{\mathrm{d}_{max}}\frac{d\mathbbm{P}(|h|^2< \zeta_{u,t} \mathrm{d}^{n_{pl}})}{d\zeta_{u,t}}f_{\mathrm{d}}(\mathrm{d} )d\mathrm{d}  \frac{d\zeta_{u,t}}{dw} 
    = \frac{\Gamma(\frac{2+n_{pl}}{n_{pl}},\zeta_{u,t}\mathrm{d}_{min}^{n_{pl}}){-}\Gamma(\frac{2+n_{pl}}{n_{pl}},\zeta_{u,t}\mathrm{d}_{max}^{n_{pl}})}   {n_{pl}\zeta_{u,t}^{ {(2+n_{pl})}/{n_{pl}} }(\mathrm{d}_{max}^2-\mathrm{d}_{min}^2)/2} \frac{d\zeta_{u,t}}{dw}.
    \end{align*}
    }.

\paragraph{Static channel ($\rho=1$)}
The channel remains the same for each retransmission. For user $u_0$, the channel is time invariant ($g_{u_0} = g_{u_0,t}\; \forall t\in [t_0,t_0+L_{u_0}-1]$) but unknown. Only the user location is known. At time $t>t_0$, we have 
\begin{eqnarray*}
\Phi({\bm{w}}_{u_0,t};\mathrm{d}_{u_0}) &=&\mathbb{P}(w_{u_0,t}\log(1{+}g_{u_0} \mathrm{P})<D_{u_0}|w_{u_0,t'}\log(1{+}g_{u_0} \mathrm{P})< D_{u_0} \forall t'\in[t_0,t{-}1], d_{u_0})\\
&=&\frac{\mathbb{P}( w_{u_0,t''}\log(1+g_{u_0} \mathrm{P})< D_{u_0}\; \forall t''\in[t_0,t]\,|\,d_{u_0})}{\mathbb{P}(w_{u_0,t'}\log(1+g_{u_0} \mathrm{P})< D_{u_0}\; \forall t'\in[t_0,t{-}1]\,|\,d_{u_0})}.
\end{eqnarray*}

Therefore, we obtain
\begin{align}
\Phi({\bm{w}}_{u_0,t};\mathrm{d}_{u_0})=
\begin{cases} \displaystyle
\frac{P_{u_0}^{fail}(\max\{ {\bm{w}}_{u_0,t}\},P ;\mathrm{d}_{u_0}  )}{P_{u_0}^{fail}(\max\{ {\bm{w}}_{u_0,t-1}\},P  ;\mathrm{d}_{u_0}  )}, \quad \text{if } t>t_0 \\
P_{u_0}^{fail}(w_{u_0,t},P ;\mathrm{d}_{u_0} ),\quad \text{if } \quad t=t_0.\\
\end{cases}
\label{eq:UserFailProbCurrent_const_current}
\end{align}
For subsequent (future) users ($u_i$ with $i\geq 1$), the expressions remain the same with the only difference that the locations of those users are also not known. Hence, in \eqref{eq:UserFailProbCurrent_const_current}, we just need to omit $\mathrm{d}_{u}$ similarly to the i.i.d. case.

\paragraph{General Markovian channel~($\rho\in(0,1)$)}
This is the most complicated case due to the correlation between channel realizations. At time $t$, the distribution of $h_{u,t}$ given the past (which is not known in practice) is Rician distributed. Specifically, if user $u$ is active at $t-1$ and $t$, we have $\mathbbm{P}(|h_{u,t}|{=}x \hspace{2pt} \Big{|} |h_{u,t-1}| )= \mathrm{Rice}(x;v_{R}=\rho|h_{u_0,t-1}|,\sigma^2_{R}=\frac{1-\rho^2}{2})$, where $v_R$ and $\sigma^2_R$ is the distance and the spread parameters respectively of the Rice distribution. 
Let us focus on user $u_0$ at time $t=t_0+1$. According to \cite[eq.(37)]{QmarcumIntegrals}, we have
\begin{eqnarray}
\Phi({\bm{w}}_{u_0,t_0+1};\mathrm{d}_{u_0}) & = & \int_{0}^{x_{u_0,0}}    \int_{0}^{x_{u_0,1}}    \mathbbm{P}(|h_{u_0,t_0+1}|{=}x \hspace{2pt} |  y ) \mathbbm{P}(|h_{u_0,t_0}|{=} y )dxdy  \nonumber \\ 
&=&1-\frac{e^{-x_1^2}Q_1(\frac{x_{u_0,0}}{\sigma_R},\frac{\rho x_{u_0,1}}{\sigma_R}) {-} e^{-x_{u_0,0}^2}Q_1(\frac{\rho x_{u_0,0}}{\sigma_R},\frac{x_{u_0,1}}{\sigma_R})}{2(1-e^{-x_{u_0,0}^2} )} 
\label{eq:UserFailProbMarkov_current}
\end{eqnarray}
with $x_{u_i,j}=\sqrt{\zeta_{u_i,t_i+j}}d^{-\frac{n_{pl}}{2}}, i\in \{0,1\}$ and $Q_M$ be the Marcum Q-function. 
            
For future users ($u_i, i\geq 1$), we have at time $t=t_i+1$ (we remind that user $u_i$ starts its transmission at time $t_i$):
\begin{align}
\Phi({\bm{w}}_{u,t_i+1})=\int_{\mathrm{d}_{min}}^{\mathrm{d}_{max}}  \Phi({\bm{w}}_{u_i,t_i+1};\mathrm{d}_{u_i}) f_\mathrm{d}(\mathrm{d})     d\mathrm{d}
\label{eq:UserFailProbMarkov_Future}
\end{align}
where $\Phi({\bm{w}}_{u_i,t_i+1};\mathrm{d}_{u_i})$ is given by \eqref{eq:UserFailProbMarkov_current} by replacing $u_0$ with $u_i$. 
Equation \eqref{eq:UserFailProbMarkov_Future} is \textit{intractable} even considering only the first two adjacent retransmissions. This is exacerbated when one considers additional transmissions. Therefore, the baseline algorithm is only designed for $\rho=0$ or $\rho=1$, even if it is also tested in the general case $\rho\in (0,1)$. Specifically, for any $\rho$, we apply the baseline algorithm designed for either $\rho=0$ or $\rho=1$ and keep the best result.

\bibliographystyle{IEEEtran}
\bibliography{IEEEabrv,Library}

\end{document}